\documentclass[a4paper,fleqn]{cas-dc}

\usepackage[numbers,sort&compress]{natbib}
\usepackage{amsmath}
\usepackage{graphicx}

\usepackage{soul}
\usepackage{bm}
\usepackage{color, xcolor}
\usepackage{subfig}
\usepackage{utfsym}
\usepackage{amssymb}
\usepackage{hyperref}

\makeatletter
\newif\if@restonecol
\makeatother

\usepackage[linesnumbered,ruled]{algorithm2e}%[ruled,vlined]{
\usepackage{algpseudocode}
\soulregister\cite7
\soulregister\citep7
\soulregister\ref7

\def\tsc#1{\csdef{#1}{\textsc{\lowercase{#1}}\xspace}}
\tsc{WGM}
\tsc{QE}
\begin{document}
\let\WriteBookmarks\relax
\def\floatpagepagefraction{1}
\def\textpagefraction{.001}
\let\printorcid\relax

\shorttitle{}
\shortauthors{}

\title [mode = title]{DFEN: Dual Feature Equalization Network  for Medical Image Segmentation}

% \tnotemark[<tnote number>] 



%\author{A}[orcid=0000-0000-0000-0000]
\author[1]{Jianjian Yin} 

% \fnmark[College of Computer Science and Engineering, Chongqing University of Technology, Chongqing 400054, China]
\ead{JianJYin_Yu@163.com}
% \ead[url]{<URL>}
% \credit{<Credit authorship details>}

% \affiliation[<aff no>]{organization={},
% %          citysep={}, % Uncomment if no comma needed between city and postcode

\author[1]{Yi Chen}   
\cormark[1]
\ead{cs_chenyi@njnu.edu.cn}

\author[1]{Chengyu Li}
\ead{evo_li@outlook.com}

\author[1]{Zhichao Zheng}
\ead{zheng_zhichaoX@163.com}

\author[1]{Yanhui Gu}
\ead{gu@njnu.edu.cn}

\author[1]{Junsheng Zhou}
\ead{zhoujs@njnu.edu.cn}

\address[1]{School of Computer and Electronic Information/School of Artiffcial Intelligence, Nanjing Normal University, China}

\cortext[1]{Corresponding author.}



\begin{abstract}
Current methods for medical image segmentation primarily focus on extracting contextual feature information from  the perspective of  the  whole  image.  While these methods have shown effective performance, none of them take into account the fact that pixels at the boundary and regions with a low number of class pixels capture more contextual feature information from other classes, leading to misclassification of pixels by unequal contextual feature information. In this paper, we propose a dual feature equalization network based on  the  hybrid architecture of Swin Transformer and Convolutional Neural Network, aiming to augment the pixel feature representations by image-level equalization feature information and class-level equalization feature information.  Firstly, the image-level feature equalization module is designed to equalize the contextual information of pixels within the image.  Secondly, we aggregate regions of the same class to equalize the pixel feature representations of the corresponding class by class-level  feature equalization module. Finally, the pixel feature representations are enhanced  by learning weights for image-level equalization feature information and class-level equalization feature information.  In addition, Swin Transformer is utilized as both the encoder and decoder, thereby bolstering the ability  of the  model to capture long-range dependencies and spatial correlations.   We conducted extensive experiments on Breast Ultrasound Images (BUSI), International Skin Imaging Collaboration (ISIC2017), Automated Cardiac Diagnosis Challenge (ACDC) and PH$^2$  datasets.  The experimental results demonstrate that our method have achieved state-of-the-art performance. Our code is publicly available at \url{https://github.com/JianJianYin/DFEN}.
\end{abstract}


\maketitle

%%%%%%%%%%%%%%%%%%%%%%%%
% \begin{table}[<options>]  
% \begin{tabular*}{\tblwidth}{@{}LL@{}}
% \toprule
% \midrule
%  & \\
%  & \\
%  & \\
%  & \\
% \bottomrule
% \end{tabular*}
% \end{table}

% Use the style of numbering in square brackets.
% If nothing is used, default style will be taken.
%\begin{enumerate}[a)]
%\item 
%\item 
%\item 
%\end{enumerate}  

%\begin{itemize}
%\item 
%\item 
%\item 
%\end{itemize}  

%\begin{description}
%\item[]
%\item[] 
%\item[] 
%\end{description}  

\section{Introduction}
\vspace{0.1in}
Medical image segmentation plays an indispensable role in medical diagnosis \cite{azad2021stacked,azad2021texture,fu2018joint, kumar2018infinet, guo2022joint, ribeiro2023left}. Its primary objective is to delineate the regions of tissue pathology within the image.  Early methods in medical image segmentation were predominantly based on edge detection and template matching 
 \cite{lee2001automated, holtzman2003hierarchical, zhan2008active}. Even though they are able to achieve exciting performance, they are still unable to meet the requirements of the application.  In recent years, deep neural networks have made remarkable strides in the field of computer vision, significantly advancing the development of medical image segmentation methods \cite{chen2022mtdcnet,li2024ucfiltransnet,xiao2023transformers}.  The UNet \cite{ronneberger2015u} architecture based on convolutional neural networks (CNNs) has greatly improved the performance of medical image segmentation and laid the foundation for future research in this field.  Many methods \cite{zhou2019unet++,gao2021utnet,milletari2016v,li2018h, huang2020unet,HAN2022109512, chen2023rethinking}  have been developed to improve upon the UNet architecture, enabling the network to incorporate a broader range of feature information. In general, methods based on the UNet architecture often employ skip connection to merge feature representations from multiple layers of the same level, allowing for communication and fusion between deep and shallow features, effectively solving the problem of detail information loss caused by reduced resolution.
However, these CNN-based methods lack the capability to capture long-range dependencies and spatial correlations. 

During the recent years,  Transformer \cite{devlin-etal-2019-bert} has achieved tremendous success in natural language processing (NLP). Shortly thereafter, Vision Transformer (Vit) \cite{dosovitskiy2021an} is proposed and utilized in image classification task, achieving higher performance than convolutional neural network (CNN).  Vit divides the image into several blocks and performs attention operations on each block, which has clear drawbacks: high computational cost and inability to be applied to tasks requiring dense predictions. Swin Transformer \cite{liu2021swin} based on Vit was proposed for dense prediction tasks and significantly reducing computational cost. In addition to  window multi-head self-attention (W-MSA) mechanism of Vit, Swin Transformer introduces shifted window multi-head self-attention (SW-MSA) to enable communication between different windows.  Several existing studies have introduced the  Transformer network architecture into medical image segmentation, showcasing many methods \cite{chen2021transunet,cao2023swin, zhang2021transfuse, 9785614, heidari2023hiformer} with impressive performance.
Some of the methods \cite{cao2023swin, 9785614} adopt a pure Swin Transformer structure, which can capture rich global features at every stage of the network training process. The other part of the methods \cite{chen2021transunet, zhang2021transfuse, heidari2023hiformer} adopt a hybrid structure of convolutional neural network and  Transformer, which combines the detailed local features generated by convolutional neural network with the global features generated by Transformer to achieve more accurate target region segmentation. It is important to note that the aforementioned methods utilizing the Transformer architecture possess the capability to capture long-range dependencies and spatial correlations.

\begin{figure*}
      \centering
      %\vspace{-0.45cm}
	\subfloat[Image]{
		\begin{minipage}[t]{0.19\textwidth}
			\centering
			\includegraphics[width=3cm,height=3cm]{image_benign_135.png}\\
			\vspace{0.03cm}
		\end{minipage}%
	}%
       \subfloat[UNeXt]{
		\begin{minipage}[t]{0.19\textwidth}
			\centering
			\includegraphics[width=3cm,height=3cm]{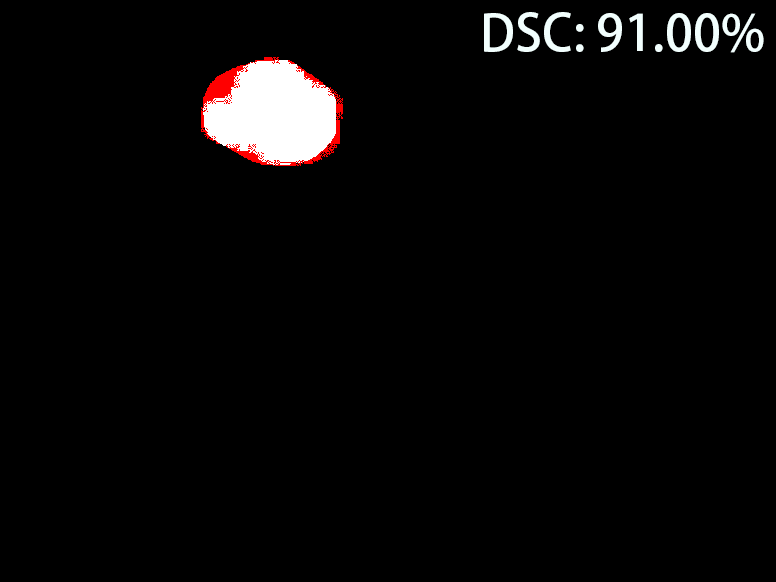}\\
			\vspace{0.03cm}
                    
		\end{minipage}%
	}%
      \subfloat[DFEN]{
		\begin{minipage}[t]{0.19\textwidth}
			\centering
			\includegraphics[width=3cm,height=3cm]{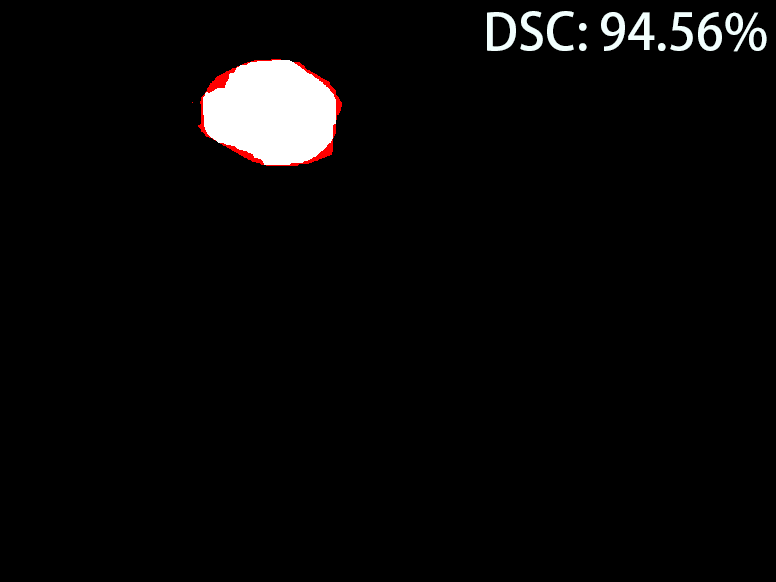}\\
			\vspace{0.03cm}
                    
		\end{minipage}%
	}%
      \subfloat[GT]{
		\begin{minipage}[t]{0.19\textwidth}
			\centering
			\includegraphics[width=3cm,height=3cm]{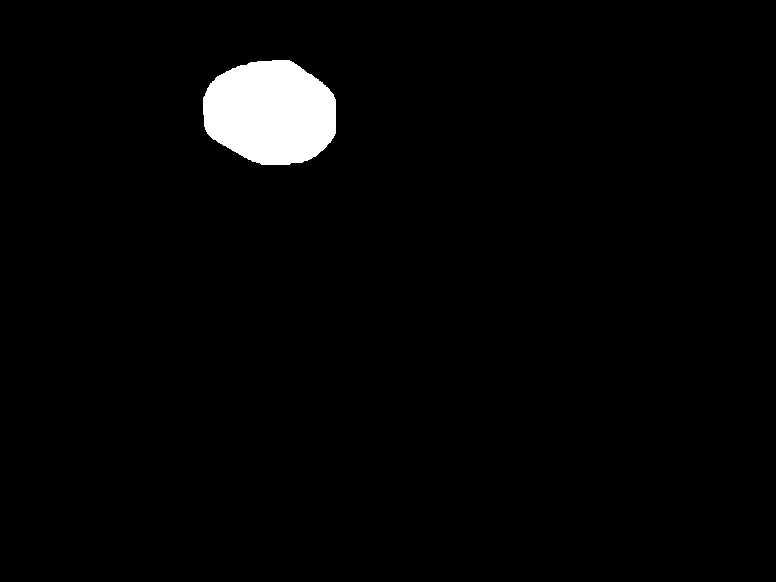}\\
			\vspace{0.03cm}
                    
		\end{minipage}%
	}%
	
	\caption{Visualization results on the BUSI dataset compared with the other state-of-the-art method UNeXt\cite{valanarasu2022unext}.  The black area show the background and the white area show the tumor. The red area indicates the area that was misclassified for each method. DSC represents the Dice Similarity Score for the entire image. GT stands for ground truth label. }
	\label{fig1}
\end{figure*}

However, both CNN and Transformer-based methods extract contextual features from the perspective of the image, ignoring the fact that pixels at the boundary and regions with fewer class pixels will capture more contextual information from other classes, which leads to misclassification of pixels.  As shown in Fig. \ref{fig1}, there are two classes in the image: background and tumor. From the image, it can be observed that the number of background pixels far exceeds that of the tumor. During the encoding process of the encoder, the tumor pixels capture a significant amount of contextual information from the background, resulting in misclassification of the tumor region pixels. Therefore, this paper proposes a dual feature equalization network (DFEN) to enhance the pixel feature representations. Firstly, the image-level feature equalization module (ILFEM) is used to equalize the contextual information of the pixels to obtain image-level equalization feature information. Next, we aggregate regional pixel features of the same class to equalize the pixel feature representations of the corresponding class to obtain the class-level equalization feature information by class-level feature  equalization module (CLFEM), with the aim of enabling the network to fully take into account the contextual information of each class. Finally, the original pixel feature representations are augmented by learning the weights of the image-level equalization feature information and the class-level equalization feature information.  In addition, we adopt the Swin Transformer architecture as the encoder and decoder, enhancing the  ability of the model to capture long-range dependencies and spatial correlations.  

In summary, our contributions are as follows: 
\begin{itemize}
    \setlength{\itemsep}{0pt}
    \setlength{\parsep}{0pt}
    \setlength{\parskip}{0pt}
    \item To the best of our knowledge, this paper is the first to enhance the pixel feature representations in the field of medical image segmentation by utilizing image-level equalization feature information and class-level equalization feature information .
    \item  The class-level feature equalization module is designed to extract the class-level equalization feature information for each class, making the network more focused on small regions of classes during training.
    \item  We design an image-level feature equalization module to extract image-level equalization feature information, which works with the class-level equalization feature information to alleviate the pixel misclassification problem caused by unequal contextual feature information.

\end{itemize}

%\subfloat[TransUNet]{
%		\begin{minipage}[t]{0.19\textwidth}
%			\centering
%			\includegraphics[width=3cm,height=3cm]{first/tranunet/benign (135).png}\\
%			\vspace{0.03cm}
                    
%		\end{minipage}%
	%}%

\section{Related  work}
\vspace{0.1in}

\subsection{CNN-based  Methods}
\vspace{0.1in}
Deep neural networks \cite{cai2024poly,yin2025semi,wu2025generative,yin2024class_class,wu2024image,li2025expert,chen2024knowledge,wuimgfu,yin2025uncertainty,chen2024spatial}  have made significant progress and development, greatly enhancing the accuracy of medical image segmentation, especially with the UNet \cite{ronneberger2015u} network architecture laying the cornerstone in medical image segmentation. Some exciting 
 CNN-based methods \cite{chen2023rethinking, fan2020pranet, 9995040, valanarasu2022unext,HAN2022109512} have enthusiastically emerged in recent years. PraNet \cite{fan2020pranet} proposes a parallel reverse attention network to accurately segment polyps in colonoscopy images. NU-Net \cite{chen2023rethinking} uses fifteen  layers of UNet to extract richer feature information, while developing a multi-output UNet as the link between the encoder and decoder to enhance the network tumor robustness  at different scales. Chained residual pooling (CRP) is designed by DDN \cite{li2018dense} to expand receptive field, and dense deconvolutional layers (DDLs) is designed to establish the relationship between neighboring pixels of the feature map to solve the ambiguous boundary shapes. DAGAN \cite{lei2020skin} constructs dense dilated convolutional blocks to enhance the transmission of effective features and protect more fine-grained structural information. MALUNet \cite{9995040} introduces four attention modules to obtain global and local information respectively, and fused information from multiple stages to generate corresponding attention maps. UNeXt \cite{valanarasu2022unext} is committed to learning good representation ability in latent space through novel tokenized  MLP blocks with axial shifts, while improving inference speed and lower computational 
 complexity. ConvUNeXt \cite{HAN2022109512} designs a lightweight attention mechanism to suppress irrelevant features, making the network more focused on the target areas.  Although CNN-based methods can achieve exciting results in tasks related to image semantic segmentation, they all suffer from two limitations. The first is that these methods tend to focus on speciffc details and struggle with global modeling due to the use of  finite-sized convolutional kernels. The second is that methods based on CNN only consider extracting contextual semantic information from the perspective of the entire image, resulting in pixels at the boundary and regions with fewer class pixels obtaining more contextual information from other classes, which leads to misclassification of pixels. The research in this paper focuses on the latter.

\begin{figure*}
\includegraphics[width=\textwidth]{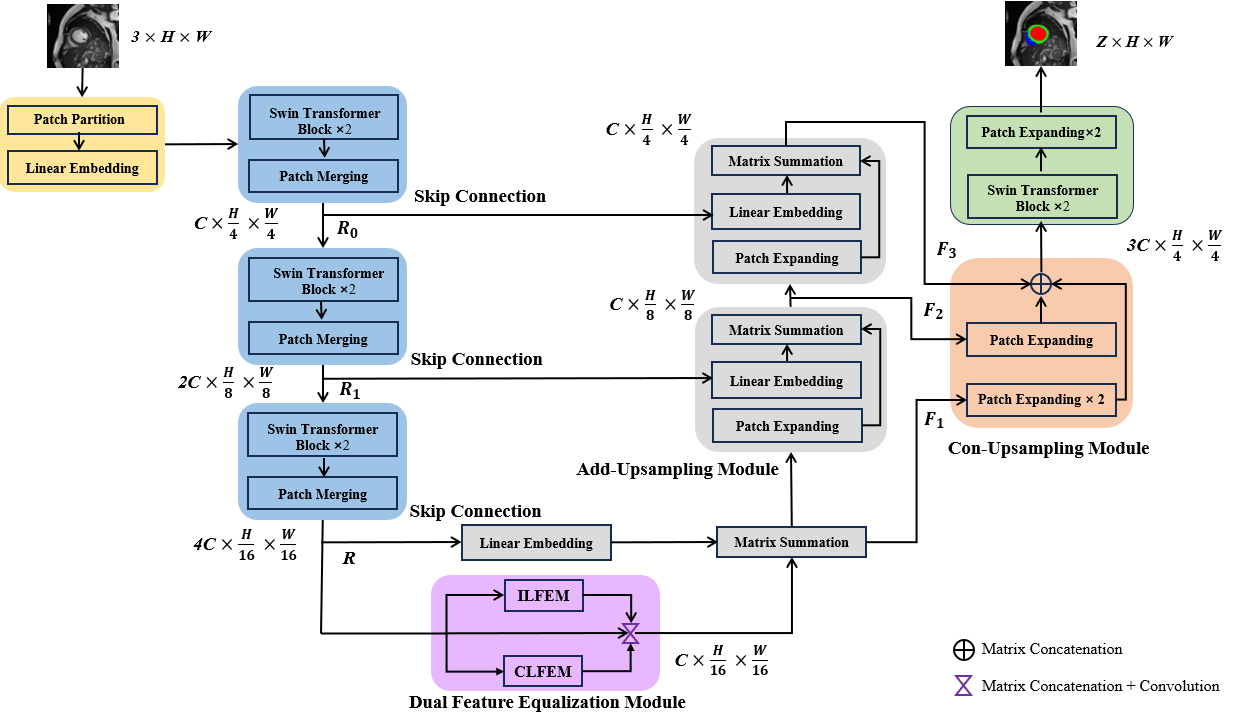}
\caption{The network structure of the DFEN model. DFEN is a hybrid framework based on Swin Transformer and CNN. Except for the dual feature equalization module, which is based on CNN, all other modules are based on Swin Transformer. The dual feature equalization module (ILFEM and CLFEM) is dedicated to enhancing pixel feature representations by utilizing class-level equalization feature information and image-level equalization feature information.
Add-Upsampling refers to additive upsampling, which is used to obtain several additive upsampling features ($F_1$, $F_2$, $F_3$) by fusing the features generated by the encoder ($R_{0}$, $R_1$, $R$) and upsampling of the corresponding depth. Similarly, Con-Upsampling is a concatenative upsampling that focuses on upsampling these additive upsampling features, and finally concatenating the channel dimensions of the features. $Z$ is the number of classes.
}
\label{fig2}
\end{figure*}

\subsection{Transformer-based Methods}
\vspace{0.1in}
In recent years, Transformer-based models \cite{dosovitskiy2021an, cao2023swin} have achieved superior performance in several tasks in computer vision, especially in medical image segmentation task. There are many Transformer-based methods\cite{hatamizadeh2022unetr, heidari2023hiformer, wang2022stepwise, park2022swine, TANG2023105131} with amazing performance in the current medical image segmentation field. HiFormer \cite{heidari2023hiformer}  proposes a novel hybrid method aimed at integrating  long-range contextual interaction information of Transformer and  local information of CNN. 
The method obtained by combining CNN-based EfffcientNet \cite{tan2019efficientnet} and Transformer is called SwinE-Net \cite{park2022swine}, which can preserve global semantic information without losing low-level semantic information. 
The CS module is proposed by TransCS-Net \cite{TANG2023105131}  to compress images into low dimensional measurements, and finally segment the target areas based on the measurements. 
The SSFormer\cite{wang2022stepwise} aims to design a PLD decoder that is well-suited for the Transformer feature pyramid. PLD decoder can effectively smooth and accentuate local features within the transformer, consequently enhancing the detailed processing capacity of neural network. UNETR\cite{hatamizadeh2022unetr} redefines three-dimensional medical image segmentation as a sequence prediction problem and introduces the architecture of UNet Transformers to learn the representations of sequences. PHCU-Net\cite{xu2023phcu}  meticulously designs a parallel hierarchical  feature extraction encoder to significantly reduce the loss of shallow texture information, thereby alleviating the problem of insufficient feature extraction in dermoscopic images. Despite the exceptional performance demonstrated by Transformer-based methods, they still fall short in effectively addressing the challenge of unequal contextual feature information.

\section{Methodology}
\vspace{0.1in}
\subsection{ Overall Architecture of the DFEN Model }
\vspace{0.1in}
   As shown in Fig. \ref{fig2}, the DFEN network primarily consists of a feature encoder with pretrained parameters on ImageNet, a dual feature  equalization module, and a feature decoder module.  The pretrained feature encoder  is formed with three consecutive downsampling modules, each module incorporating swin transformer block × 2 and patch merging. In contrast, the feature decoder consists of two additive upsampling modules, one concatenative upsampling module and an upsampling block (containing swin transformer block × 2 and patch expanding × 2). Patch expanding is used to upsample images to increase their resolution, while line embedding is dedicated to changing the channel dimension of feature maps.  The feature encoder utilizes multiple Swin Transformer layers to generate features at different levels. The deepest layer feature is fed into the dual feature equalization module. The image-level feature equalization module (ILFEM) equalizes the contextual information captured by each pixel from the perspective of the entire image to get image-level equalization feature information.  The class-level feature equalization module (CLFEM) equalizes the pixel feature representations for each class to get class-level equalization feature information.  By combining the image-level equalization feature information and class-level equalization feature information through learned weights, the pixel feature representations are enhanced, allowing the network to consider the features of each pixel more comprehensively. The feature decoder module decodes the enhanced feature and the features generated by the encoder at different levels to obtain the final prediction result.  It is important to note here that for the purpose of simplifying the  complexity of the model, we perform element-wise addition instead of concatenating the channel dimension between the encoder features obtained through skip connection and the corresponding decoder-generated features. 

\begin{figure}[h]
\centering
\includegraphics[width=0.48\textwidth,height=1.8in]{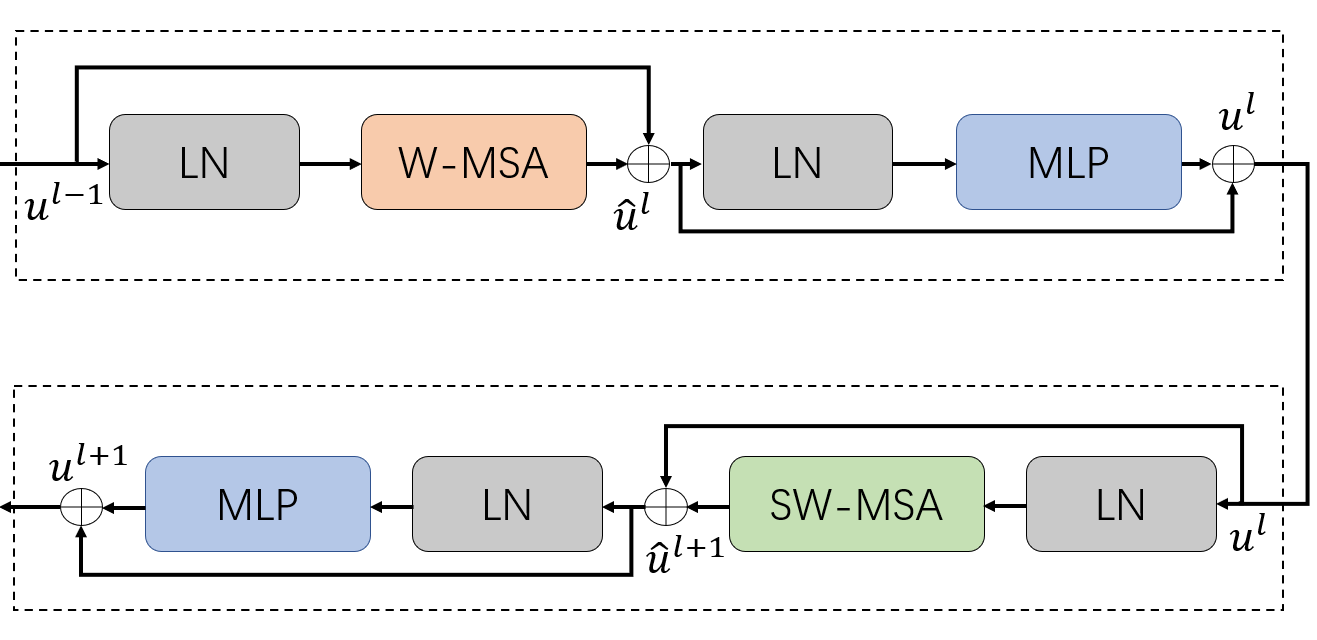}
\caption{The internal structure diagram of the Swin Transformer Block.  The Swin Transformer mainly consists of layer normalization(LN), window-based multi-head self-attention(W-MSA), shifted window-based multi-head self-attention(SW-MSA), and multi-layer perceptron(MLP).} 
\label{fig3}
\end{figure}

\subsection{Swin Transformer Block}
\vspace{0.1in}
The internal structure of the Swin Transformer Block is shown in Fig. \ref{fig3}. Unlike the Vision Transformer, the Swin Transformer proposes a SW-MSA that allows multiple windows to communicate with each other, while also significantly reducing the complexity of the computation.   The principle of the Swin Transformer block is described by the following consecutive equations:
\begin{equation}
\begin{aligned}
\hat{u}^{l} & =\mathrm{W}-\mathrm{MSA}\left(\mathrm{LN}\left(u^{l-1}\right)\right)+u^{l-1} \\
u^{l} & =\operatorname{MLP}\left(\mathrm{LN}\left(\hat{u}^{l}\right)\right)+\hat{u}^{l} \\
\hat{u}^{l+1} & =\operatorname{SW}-\mathrm{MSA}\left(\mathrm{LN}\left(u^{l}\right)\right)+u^{l} \\
u^{l+1} & =\operatorname{MLP}\left(\mathrm{LN}\left(\hat{u}^{l+1}\right)\right)+\hat{u}^{l+1},
\end{aligned}
\label{eq1}
\end{equation}
Following \cite{hu2018relation,hu2019local}, the attention can be computed using the formula provided below:
\begin{equation}
\begin{aligned}
\operatorname{Attention}(Q, K , V)=\operatorname{SoftMax}\left(\frac{Q K^{T}}{\sqrt{d}}+B\right) V
\end{aligned}
\label{eq2}
\end{equation}
$Q$/$K$/$V$ respectively stand for query/key/value matrix, $B$ represents the bias matrix, while $d$ refers to the number of channels.

\begin{algorithm}[t]
\caption{DFEN algorithm in a mini-batch}
\KwIn{$ {B_l} = \{ ({I_i},{y_i})\} _{i = 1}^{|{B_l}|} $ : Mini-batch Images;}
\KwOut{ $\mathcal{L}_{loss}$ : Training loss for updating network; }
\textbf{Initialize}:  \\
\hspace{0.3cm}  encoder $\varphi $, decoder $d$, loss weight($\alpha$,$\beta$), \\
\hspace{0.3cm}  image-level feature equalization module $ \vartheta $,  \\
\hspace{0.3cm}   class-level feature equalization module $ \theta $, \\
\hspace{0.3cm}  feature fusion operation $\phi$;
\\

\Begin
{
\For{each $I_i$ $\in$ $ B_l$}
{
${R_0},{R_1},R = \varphi  ({I_i})$;\\
\textcolor{blue}{\footnotesize \# get image-level equalization feature representations }\\
$R_{il}$ = $\vartheta$($R$);\\
\textcolor{blue}{\footnotesize \# get class-level equalization feature representations }\\
$R_{cl}$ = $\theta$($R$);\\
\textcolor{blue}{\footnotesize \# feature augmentation}\\
$R_{aug}$ = $\phi ({R},{R_{il}},{R_{cl}})$; \\
 \textcolor{blue}{\footnotesize \# get predicted result} \\
$p$  =  $d(R_{0},R_{1},R,R_{aug})$; \\
\textcolor{blue}{\footnotesize \# compute loss} \\
$\mathcal{L}_{loss}$ = $\alpha$ × $\mathcal{L}_{ce}$($p$,$y_i$) + $\beta$ × $\mathcal{L}_{dice}$($p$,$y_i$); \\
}
\textbf{return:} avg($\mathcal{L}_{loss}$).
}
\end{algorithm}

\subsection{Image-level Feature Equalization Module}
\begin{figure}
\centering
\includegraphics[width=0.45\textwidth,height=4.4cm]{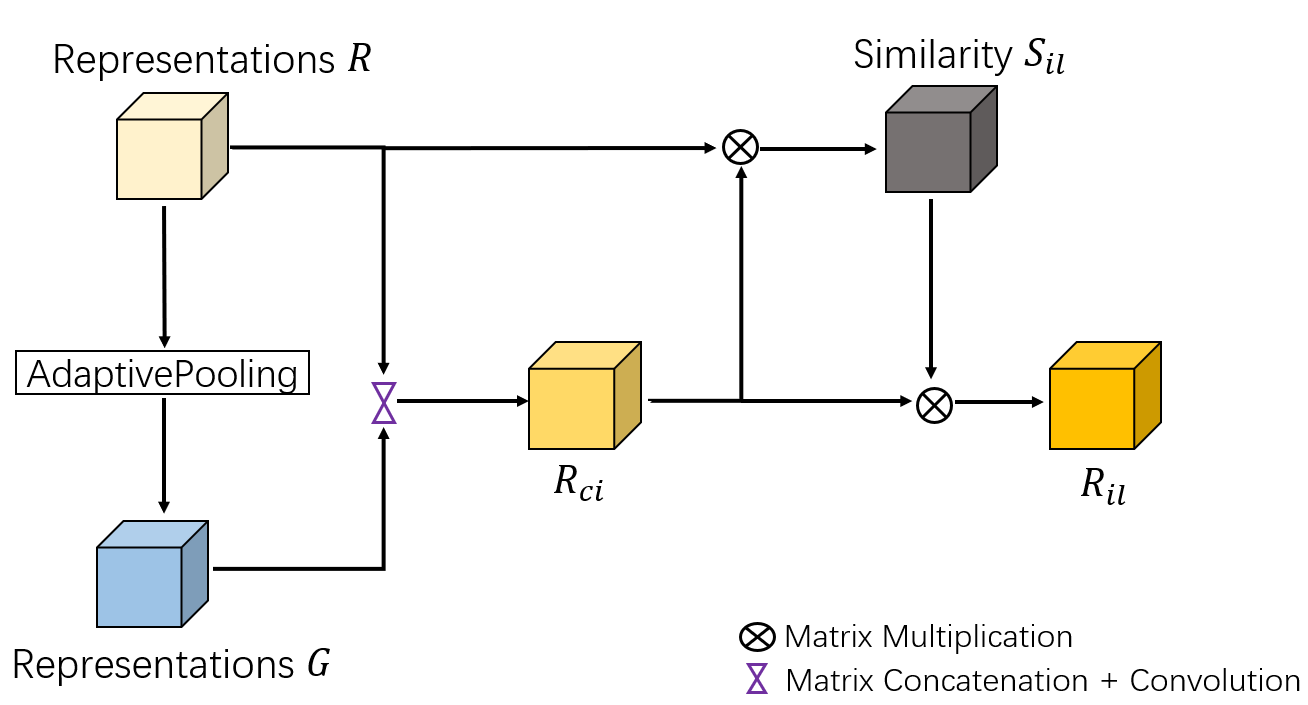}
\caption{The overview diagram of the image-level feature equalization module. $R_{ci}$ is the coarse image-level equalization feature representations. and $R_{il}$ is the fine-grained image-level equalization feature representations.}
\label{fig4}
\end{figure}

\vspace{0.1in}
 Existing medical image segmentation methods extract contextual information from the entire image, neglecting  the fact that boundary pixels and regions with fewer class pixels capture more contextual information from other classes.  Therefore, an image-level feature equalization module is used to equalize the contextual information captured by each pixel to prevent the  network from ignoring information from classes with a small number of pixels or boundary pixels. The structure of the image-level feature equalization module is illustrated in Fig. \ref{fig4}.

 Given an image $I$ with the size of  $ 3 \times H \times W$.   $H$ denotes the height of the image, while $W$ represents its width.   The encoder $\varphi $ generates feature representations ($R_{0},R_{1},R$) at various depths:
\begin{equation}
\begin{aligned}
R_{0},R_{1},R = \varphi  (I)
\end{aligned}
\label{eq3}
\end{equation}
 the size of $R$ is $C{\rm{ \times }}\frac{H}{{16}}{\rm{ \times }}\frac{W}{{16}}$ . $C$ indicates the number of channels. We perform global adaptive pooling($GAP$)  on the pixel feature representations $R$ to obtain the global feature representations $G$:
\begin{equation}
\begin{aligned}
G = GAP(R) \\
\end{aligned}
\label{eq4}
\end{equation}
where $G$ is a matrix of size $C{\rm{ \times }}1{\rm{ \times }}1$. It is important to note here that global adaptive pooling uses softmax to get the corresponding weights for weighted aggregation, rather than simple averaging in global average pooling. Then we obtain coarse image-level equalization feature representations ${R_{ci}}$ based on the pixel representations $R$ and the global pixel representations $G$:
\begin{equation}
\begin{aligned}
{R_{ci}} = \phi  (R,upsample(G))\\
\end{aligned}
\label{eq5}
\end{equation}
 where $R_{ci}$ is a matrix of size $C{\rm{ \times }}\frac{H}{{16}}{\rm{ \times }}\frac{W}{{16}}$.  $\phi$ represents the concatenation and convolution operations. $upsample$ denotes the operation of upsampling $G$ to match the size of $R$.  We calculate the similarity ${S_{il}}$ between $R$ and ${R_{ci}}$ to refine the coarse image-level equalization feature representations ${R_{ci}}$:
\begin{equation}
\begin{aligned}
{S_{il}} = Soft\max (\frac{{{R^{\frac{{HW}}{{256}}{\rm{ \times }}C}} \otimes R_{ci}^{C{\rm{ \times }}\frac{{HW}}{{256}}}}}{{\sqrt C }}) \\
\end{aligned}
\label{eq6}
\end{equation}
where $S_{il}$ is a matrix of size $\frac{{HW}}{{256}}{\rm{ \times }}\frac{{HW}}{{256}}$. $\otimes$ stands for matrix multiplication. Finally, we obtain fine-grained image-level equalization feature representations ${R_{il}}$:
\begin{equation}
\begin{aligned}
{R_{{il}}} = resize(S_{il}^{\frac{{HW}}{{256}}{\rm{ \times }}\frac{{HW}}{{256}}} \otimes R_{ci}^{\frac{{HW}}{{256}}{\rm{ \times }}C}) \\
\end{aligned}
\label{eq7}
\end{equation}
 where $resize$ serves to adjust the dimensions of matrix $R_{{il}}$ to become $C{\rm{ \times }}\frac{H}{{16}}{\rm{ \times }}\frac{W}{{16}}$.

\begin{figure}
\centering
\includegraphics[width=0.45\textwidth]{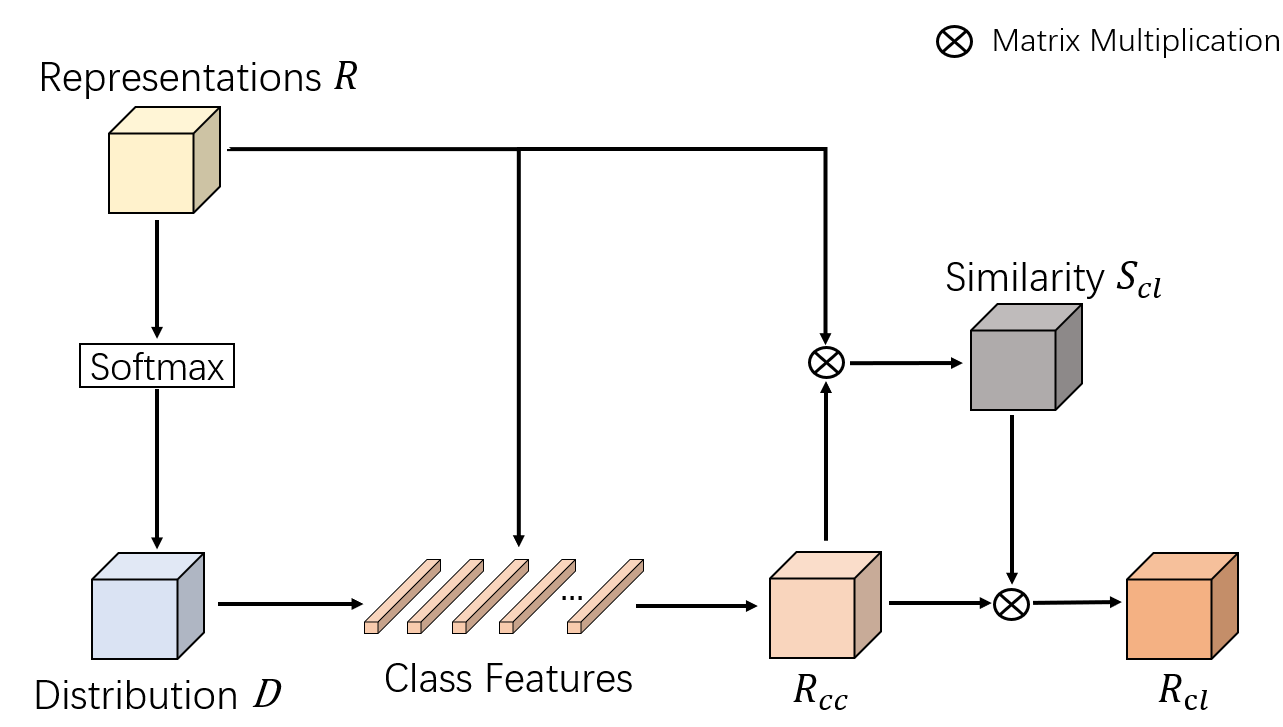}
\caption{The overview diagram of the class-level feature equalization module. $R_{cc}$  is the  coarse class-level equalization feature representations, and $R_{cl}$  is the fine-grained class-level equalization feature representations.} 
\label{fig5}
\end{figure}

\subsection{Class-level Feature Equalization Module}
\vspace{0.1in}
The class-level feature equalization module aggregates regions of the same class to equalize the pixel feature representations of the corresponding class. The specific structure is shown in Fig.  \ref{fig5}.

  We apply softmax to the pixel feature representations $R$ to obtain the probability distribution $D$.  The size of $D$ is $ Z{\rm{ \times }}\frac{H}{{16}}{\rm{ \times }}\frac{W}{{16}}$, and $Z$ is the number of classes. Based on $D$, the pixel feature representations $R$ is divided into several class regions:
\begin{equation}
\begin{aligned}
{F_c} = \{ {R_{[*,i,j]}}|\arg \max ({D_{[*,i,j]}}) = c\} 
\end{aligned}
\label{eq8}
\end{equation}
where $i$ and $j$ represent the positional coordinates on the  feature representations $R$. The size of $F_c$ is ${N_c}{\rm{ \times }} C$, and ${N_c}$ represents the number of features belonging to class c.   Correspondingly, we obtain the class probability distribution values using the following formula:
\begin{equation}
\begin{aligned}
{D_{\rm{c}}} = \{ {D_{[c,i,j]}}|\arg \max ({D_{[*,i,j]}}) = c\} 
\end{aligned}
\label{eq9}
\end{equation}
we obtain the corresponding class feature representations after weighted aggregation based on $D$ and $F_c$:
\begin{equation}
\begin{aligned}
R_c = \sum\limits_{i = 1}^{{N_c}} {\frac{{{e^{{D_{c,[i,*]}}}}}}{{\sum {{e^{{D_c}}}} }}} {F_{c,[i,*]}}
\end{aligned}
\label{eq10}
\end{equation}
where $R_c$ denotes the feature representation of class c with dimensions $1 {\times} C$ . The following formula is used to aggregate the features of each class into the basic pixel feature representations $R$, resulting in the coarse class-level equalization feature representations $R_{cc}$:
\begin{equation}
\begin{aligned}
{R_{cc,[*,i,j]}} = {R_c}\;\;if\,\arg \max ({D_{[*,i,j]}}) = c
\end{aligned}
\label{eq11}
\end{equation}
 where $R_{cc}$ is a matrix of size $C{\rm{ \times }}\frac{H}{{16}}{\rm{ \times }}\frac{W}{{16}}$ . Next, we calculate the similarity $S_{cl}$ between $R$ and $R_{cc}$ to refine the coarse class-level equalization feature representations $R_{cc}$:
\begin{equation}
\begin{aligned}
{S_{cl}} = Soft\max (\frac{{{R^{\frac{{HW}}{{256}}{\rm{ \times }}C}} \otimes R_{cc}^{C{\rm{ \times }}\frac{{HW}}{{256}}}}}{{\sqrt C }}) \\
\end{aligned}
\label{eq12}
\end{equation}
where $S_{cl}$ is a matrix of size $\frac{{HW}}{{256}}{\rm{ \times }}\frac{{HW}}{{256}}$. Finally,
the fine-grained class-level equalization  feature representations $R_{cl}$ are obtained by following formula:
\begin{equation}
\begin{aligned}
{R_{{cl}}} = resize(S_{cl}^{\frac{{HW}}{{256}}{\rm{ \times }}\frac{{HW}}{{256}}} \otimes R_{cc}^{\frac{{HW}}{{256}}{\rm{ \times }}C}) \\
\end{aligned}
\label{eq13}
\end{equation}
the size of $R_{cl}$ is $C{\rm{ \times }}\frac{H}{{16}}{\rm{ \times }}\frac{W}{{16}}$ . The pixel feature representations $R$ are enhanced by combining the image-level equalization feature representations  $R_{il}$ with the class-level equalization feature representations $R_{cl}$:
\begin{equation}
\begin{aligned}
{R_{aug}} = \phi ({R},{R_{il}},{R_{cl}})
\end{aligned}
\label{eq14}
\end{equation}
 where $R_{aug}$ has the same size as $R_{cl}$.

\subsection{Additive Upsampling Module}
\vspace{0.1in}
The features produced by the deeper layers of the neural network are rich in semantic information, while the features produced by the shallow network contain rich detail features. Based on the above point, as shown in Fig. \ref{fig2}, the additive upsampling module utilizes the features generated by the encoder at different depths obtained by skip connection and fuses them with the features generated by upsampling to obtain several additive upsampling features ($F_1$, $F_2$, $F_3$). These additive upsampling features contain feature information at different scales.
The ablation experiment in the experimental section proved the effectiveness of additive upsampling module.

\subsection{Concatenative Upsampling Module}
In order to enable the network to fully consider feature information at different scales, we adopted concatenative upsampling module to upsample several additive upsampling features ($F_1$, $F_2$, $F_3$), and concatenated them in the channel dimension. The ablation experiment has demonstrated that the concatenative upsampling module can improve the performance of the model.

\section{Experiments}
\vspace{0.1in}
\subsection{Datasets $\& $ Metric}
\vspace{0.1in}
\begin{sloppypar}
We have selected the following datasets for experimentation to demonstrate the superiority of the proposed method:
Automated Cardiac Diagnosis Challenge (ACDC) \cite{bernard2018deep},
International Skin Imaging Collaboration (ISIC2017) \cite{codella2018skin}, PH$^2$ \cite{mendoncca2013ph}, Breast Ultrasound Images (BUSI) \cite{al2020dataset}.  
\end{sloppypar}

ACDC \cite{bernard2018deep} is a cardiac MRI dataset primarily focused on left ventricle(LV), right ventricle(RV) and myocardium(Myo) segmentation. The dataset is divided into 70 training images, 10 validation images and 20 test images. Same settings as ISIC2017 and BUSI dataset, ACDC images are uniformly set to 224×224.

ISIC2017 \cite{codella2018skin} is a large binary classification dataset for skin cancer segmentation containing a total of 2750 images. 2000 images were used for training, part of the 150 images used as the validation set. and the remaining 600 images were used as the testing set. We crop all the images to 224×224.

PH$^2$ \cite{mendoncca2013ph} is a skin cancer segmentation dataset, the size of the input image is uniformly set to 224 x 224. We use the same data partitioning as HiFormer\cite{heidari2023hiformer}.

The BUSI \cite{al2020dataset} dataset is dedicated to the segmentation of breast cancer. We use the data partitioning  of  \cite{chen2021transunet,  zhang2021transfuse}  to divide the dataset into training and testing sets.  BUSI images are uniformly cropped to 224×224.

We adopt specific metrics for the given task to ensure fairness. The metrics primarily include: (1) Dice Similarity Score(DSC), (2) Sensitivity(SE), (3) Specificity(SP), (4) Accuracy(ACC). An important aspect to note is that we followed the same experimental metric configurations of other state-of-the-art methods on specific datasets, instead of uniformly adopting all four metrics. 

\begin{table}[h]
\centering
\caption{Comparison of results with State-of-the-art methods on the ACDC testing set.  Similar to other methods, we used Dice Similarity Score(DSC) for three classes(RV, Myo, LV) and the average DSC  to assess the performance of the model.}
\label{tab1}
\resizebox{\linewidth}{!}{
\begin{tabular}{cc|ccc}
\hline
\hline
\textbf{Method}	& \textbf{DSC(\%)}&	\textbf{RV}&\textbf{Myo}  &\textbf{LV}\\
\hline
R50 UNet\cite{chen2021transunet} & 87.60 & 84.62 & 84.52 & 93.68 \\
R50 AttnUNet\cite{chen2021transunet} & 86.90 & 83.27 & 84.33 & 93.53 \\
ViT-CUP\cite{chen2021transunet} & 83.41 & 80.93 & 78.12 &91.17 \\
R50 ViT\cite{chen2021transunet} & 86.19 & 82.51 & 83.01 & 93.05 \\
TransUNet\cite{chen2021transunet}  & 89.71& 86.67& 87.27 & 95.18 \\
SwinUNet\cite{cao2023swin} & 88.07&  85.77 & 84.42& 94.03 \\
UNeXt\cite{valanarasu2022unext} & 89.24 & 86.76 &  86.28 & 94.69 \\
UNETR\cite{hatamizadeh2022unetr} & 88.61 & 85.29 & 86.52 & 94.02 \\
HiFormer\cite{heidari2023hiformer} & 89.68 & 87.49 & 86.55 & 94.99 \\
\hline
DFEN & \textbf{90.46} & \textbf{88.41} & \textbf{87.81} & 95.17\\
\hline
\hline
\end{tabular}
}
\end{table}

\begin{table}[h]
\centering
\caption{Comparison of results with State-of-the-art methods on the ISIC2017 testing set. For the sake of fairness, we employed four metrics to assess the  effectiveness of the model. The mark “-” shows the corresponding information is not publicly available.}
\label{tab2}
\resizebox{\linewidth}{!}{
\begin{tabular}{ccccc}
\hline
\hline
\textbf{Method}	& \textbf{DSC(\%)}&	\textbf{SE(\%)}&\textbf{SP(\%)}  &\textbf{ACC(\%)}\\
\hline
UNet\cite{ronneberger2015u}   &  81.59 & 81.72 & 96.80 &91.64 \\
UNet++\cite{zhou2019unet++}  & 85.80 & - & - & 93.80 \\
Att-UNet\cite{oktay2018attention} & 80.82 &79.98 & 97.76 &91.45 \\
DAGAN\cite{lei2020skin}  & 84.25 & 83.63 & 97.16 & 93.04 \\
TransUNet\cite{chen2021transunet}  & 81.23 & 82.63 &  95.77 &92.07 \\
MedT\cite{valanarasu2021medical}  & 80.37 & 80.64 & 95.46 & 90.90 \\
DDN\cite{li2018dense} & 86.60& - &- &93.90\\
FrCN\cite{al2018skin} & 87.10 & - &- &94.00\\
FAT-Net\cite{wu2022fat}  & 85.00 & 83.92 & 97.25 & 93.26 \\
TransFuse\cite{zhang2021transfuse} & 87.20 &- & - &94.40 \\
PraNet\cite{fan2020pranet} & 88.67 & 90.31 & 96.86 & 95.74 \\

SSFormer\cite{wang2022stepwise} & 89.48 & 90.62 & 97.24 & 96.10 \\

PHCU-Net\cite{xu2023phcu} & 89.48 & 87.72 & \textbf{98.21} & \textbf{96.41} \\
\hline
DFEN & \textbf{90.13} &\textbf{91.03} &97.50& 96.39\\
\hline
\hline
\end{tabular}
}
\end{table}

\subsection{Implementation Details}
\vspace{0.1in}
 Our model is implemented using the PyTorch framework and the experiments were conducted on a NVIDIA 3090 GPU.  The learning rate for the ISIC2017 dataset is set to 0.05, with the batch size of 24  and it is trained for 200 epochs using the stochastic gradient descent algorithm. For the BUSI dataset and ACDC dataset, the learning rate is set to 0.0001, with batch size of 8 and 24 respectively,  they are trained for 200 epochs using the Adam algorithm. The experimental setup for the PH$^2$ dataset is consistent with that of the ISIC2017 dataset. We train the network jointly using Dice loss and  Cross-entropy  loss:
\begin{equation}
\begin{aligned}
\mathcal{L}_{\text {Loss }}= \alpha \mathcal{L}_{\text {Cross-entropy loss}} +  \beta \mathcal{L}_{\text {Dice loss}}
\end{aligned}
\end{equation}
We set $\alpha$ to be 0.3 and $\beta$ to be 0.7.  We conducted detailed ablation experiments on $\alpha$ and $\beta$, the experimental results are presented in the following section.

\subsection{Comparison with  State-of-the-art Methods}
\vspace{0.1in}
    
\subsubsection{Results of ACDC dataset} 
\vspace{0.1in}
The experimental results on the ACDC dataset are presented in Table \ref{tab1}. DFEN has demonstrated superior performance in terms of DSC metrics, surpassing  UNeXt \cite{valanarasu2022unext} and SwinUNet \cite{cao2023swin} by 1.22\% and 2.39\% respectively. Additionally, DFEN has outperformed UNeXt \cite{valanarasu2022unext} by 1.65\%, 1.53\%, and 0.48\% in RV, Myo, and LV segmentation results respectively.
Our method outperforms the current state-of-the-art method HiFormer \cite{heidari2023hiformer} by 0.78\% on DSC.
The data in the table demonstrates that DFEN attains state-of-the-art result in terms of the DSC metric. We conducted detailed segmentation result visualization experiments on the ACDC dataset in the subsequent experimental section.

\subsubsection{Results of ISIC2017 dataset}   
The results of the experiments on the ISIC 2017 dataset are shown in Table \ref{tab2}.  It is clear to see that DFEN outperforms the PraNet \cite{fan2020pranet} and SSFormer \cite{wang2022stepwise} methods in all metrics, especially in the DSC metric by 1.46\% and 0.65\% respectively. In addition to this, our method clearly outperforms PHCU-Net \cite{xu2023phcu} on the DSC and SE metrics  by 0.65\%, 3.31\%, respectively, thus demonstrating that DFEN achieves state-of-the-art results.

\subsubsection{Results of PH$^2$ dataset} 
Table \ref{tab3} shows the performance comparison of our method with other state-of-the-art methods on the PH$^2$ dataset. From the experimental results in the table, it can be seen that our method achieved the best performance in DSC, SP and ACC metrics. Compared with the TransCS-Net \cite{TANG2023105131} method, our method outperforms 0.3\%, 2.28\%, and 0.72\% in three metrics, respectively. In addition, our method also outperforms the state-of-the-art method HiFormer \cite{heidari2023hiformer} in DSC, SP and ACC metrics by 0.31\%, 0.83\%, and 0.13\%, respectively.    The above experimental results demonstrate the superiority of our method for  skin cancer segmentation task.

\subsubsection{Results of  BUSI dataset} 

The experimental results on the BUSI dataset are displayed in Table \ref{tab4}. DFEN outperforms the state-of-the-art methods PraNet \cite{fan2020pranet}, 
 SSFormer \cite{wang2022stepwise}, HiFormer \cite{heidari2023hiformer} by 1.78\%, 0.96\%  and 0.43\%, respectively, on the DSC metric, respectively.  Our method outperforms state-of-the-art methods in experimental results on four datasets because it extracts image-level and class-level equalization feature information to enhance pixel feature representations, thereby alleviating pixel misclassification problem caused by unequal contextual information.

\begin{table}[h]
\centering
\caption{Comparison of results with State-of-the-art methods on the PH$^2$ testing set. For the sake of fairness, we employed four metrics to measure  the effectiveness of the model.}
\label{tab3}
\resizebox{\linewidth}{!}{
\begin{tabular}{ccccc}
\hline
\hline
\textbf{Method}	& \textbf{DSC(\%)}&	\textbf{SE(\%)}&\textbf{SP(\%)}  &\textbf{ACC(\%)}\\
\hline
UNet\cite{ronneberger2015u}   & 89.36  & 91.25 & 95.88 & 92.33 \\
Att-UNet\cite{oktay2018attention} & 90.03 & 92.05 & 96.40  & 92.76 \\
DAGAN\cite{lei2020skin}  & 92.01  & 83.20 & 96.40  & 94.25 \\
%TransUNet\cite{chen2021transunet}  & 88.40 & 90.63 & 94.27  & 92.00 \\
MedT\cite{valanarasu2021medical}  & 91.22 & 84.72 & 96.57 & 94.16  \\
% FAT-Net\cite{wu2022fat}  & 94.40 & 94.41 & 97.41 & 97.03 \\
% \textcolor{red}{TMU-Net}\cite{azad2022contextual} &  94.14 & 93.95 & 97.56 & 96.47 \\
% SwinUNet\cite{cao2023swin} & 94.49 &  94.10  & 95.64 & 96.78 \\
HorNet\cite{NEURIPS2022_436d042b} & 88.94 & 95.67 & 90.73 & 92.32 \\
MALUNet\cite{9995040} & 83.30 & 84.77 & 95.26 & 93.14 \\

MHorUNet\cite{WU2024105517} & 91.98 & 94.98 & 94.51 & 94.66 \\
HiFormer\cite{heidari2023hiformer} & 94.51 & 95.61 & 96.91 & 96.59 \\
TransCS-Net\cite{TANG2023105131} & 94.52 & 94.81 & 95.46 & 96.00 \\
GREnet\cite{wang2023grenet} & 93.50 & \textbf{95.80} & 94.20 &  96.10 \\
\hline
DFEN & \textbf{94.82} & 94.38 & \textbf{97.74} & \textbf{96.72}\\
\hline
\hline
\end{tabular}
}
\end{table}

\begin{table}[h]
\caption{Comparison of results with State-of-the-art methods on the BUSI testing set. }
\label{tab4}
\centering

\begin{tabular}{cc}
\hline
\hline
\textbf{Method}	& \textbf{DSC(\%)}\\
\hline
UNet\cite{ronneberger2015u}   &  76.35 \\
UNet++\cite{zhou2019unet++}  & 77.54 \\
ResUNet\cite{zhang2018road} & 78.25 \\
MeT\cite{valanarasu2021medical} & 76.93 \\
PraNet\cite{fan2020pranet} & 78.44 \\
TransUNet\cite{chen2021transunet}  & 79.30 \\
UNeXt\cite{valanarasu2022unext} & 79.37 \\
SSFormer\cite{wang2022stepwise} & 79.26 \\
NU-Net\cite{chen2023rethinking} & 79.42\\
HiFormer\cite{heidari2023hiformer} & 79.79 \\
\hline
DFEN & \textbf{80.22} \\
\hline
\hline
\end{tabular}

\end{table}

\subsubsection{Comparison of visualization results with State-of-the-art methods.}
\begin{figure*}
      \centering
	\subfloat[Image]{
		\begin{minipage}[t]{0.18\textwidth}
			\centering
			\includegraphics[width=3.16cm,height=2.9cm]{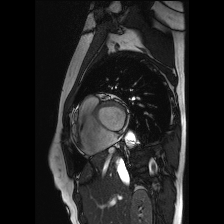}\\
			\vspace{0.03cm}
       
                       \includegraphics[width=3.16cm,height=2.9cm]{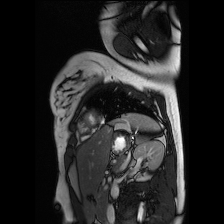}\\
			\vspace{0.03cm}
                         \includegraphics[width=3.16cm,height=2.9cm]{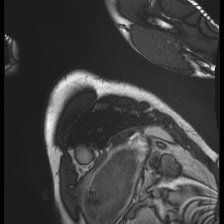}\\
			\vspace{0.03cm}
		\end{minipage}%
	}%
	\subfloat[UNeXt]{
		\begin{minipage}[t]{0.18\textwidth}
			\centering
			\includegraphics[width=3.16cm,height=2.9cm]{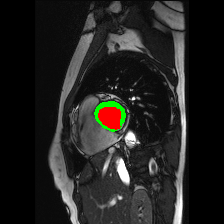}\\
			\vspace{0.03cm}
                   
                        \includegraphics[width=3.16cm,height=2.9cm]{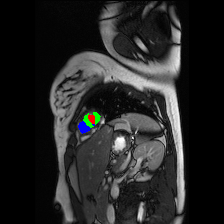}\\
			\vspace{0.03cm}
                        
                       \includegraphics[width=3.16cm,height=2.9cm]{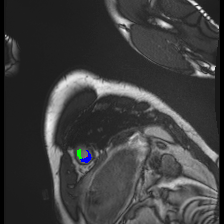}\\
			\vspace{0.03cm}
		\end{minipage}%
	}%
       \subfloat[SwinUNet]{
		\begin{minipage}[t]{0.18\textwidth}
			\centering
			\includegraphics[width=3.16cm,height=2.9cm]{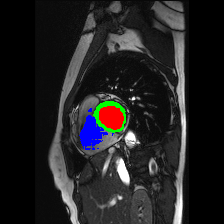}\\
			\vspace{0.03cm}
                    
                        \includegraphics[width=3.16cm,height=2.9cm]{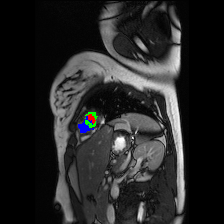}\\
			\vspace{0.03cm}
                        \includegraphics[width=3.16cm,height=2.9cm]{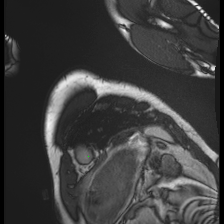}\\
			\vspace{0.03cm}
                    
		\end{minipage}%
	}%
      \subfloat[DFEN]{
		\begin{minipage}[t]{0.18\textwidth}
			\centering
			\includegraphics[width=3.16cm,height=2.9cm]{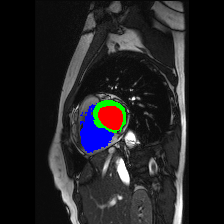}\\
			\vspace{0.03cm}
                   
                        \includegraphics[width=3.16cm,height=2.9cm]{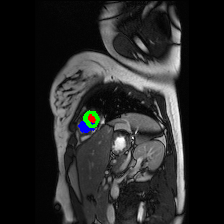}\\
			\vspace{0.03cm}
                        \includegraphics[width=3.16cm,height=2.9cm]{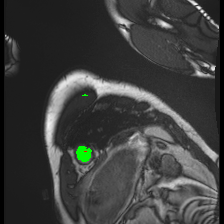}\\
			\vspace{0.03cm}
		\end{minipage}%
	}%
      \subfloat[GT]{
		\begin{minipage}[t]{0.18\textwidth}
			\centering
			\includegraphics[width=3.16cm,height=2.9cm]{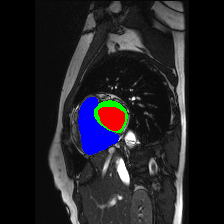}\\
			\vspace{0.03cm}
                     
                        \includegraphics[width=3.16cm,height=2.9cm]{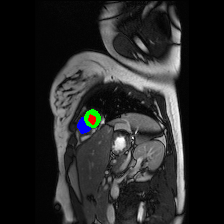}\\
			\vspace{0.03cm}

                        \includegraphics[width=3.16cm,height=2.9cm]{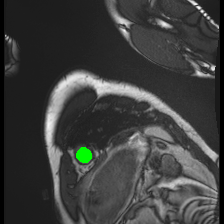}\\
			\vspace{0.03cm}
                    
		\end{minipage}%
	}%
	
	\caption{Qualitative results with other state-of-the-art methods in the ACDC dataset.  The red region represents the LV, the green region indicates the  Myo, and the blue region corresponds to the RV.}
	\label{fig6}
\end{figure*}

\vspace{0.1in}
In Fig. \ref{fig6}, we showcase the qualitative comparison between the results of DFEN and those achieved by other state-of-the-art approaches on the ACDC dataset.  In the first row, the RV segmented by DFEN is clearly more accurate than the other methods.  The results in the second and third rows illustrate respectively that DFEN can alleviate the misclassification problem caused by regions with few class pixels and boundary pixels being able to capture more contextual information from other classes.

\begin{table}
\caption{Comparison of model complexity with other state-of-the-art (SOTA) methods. A smaller value indicates a more lightweight model.}
\label{tab5}
\centering
\begin{tabular}{cc}
\hline
\hline
\textbf{Method}	& \textbf{Params(M)}\\
\hline
PraNet\cite{fan2020pranet} & 32.55 \\
SSFormer\cite{wang2022stepwise} & 66.22 \\
TransUNet \cite{chen2021transunet} & 105.32\\
ResUNet\cite{zhang2018road} & 62.74 \\
UNet\cite{ronneberger2015u}& 31.13 \\
HiFormer\cite{heidari2023hiformer} & 29.52 \\
SwinUNet\cite{cao2023swin} & 27.17 \\
\hline
DFEN & \textbf{24.01} \\
\hline
\hline
\end{tabular}

\end{table}

\begin{table}
\caption{Ablation experiments  to assess the effectiveness of ILFEM and CLFEM on the BUSI testing set.}
\label{tab6}
\centering
\begin{tabular}{cccc}
\hline
\hline
\textbf{Baseline} &  \makebox[0.12\textwidth][c]{\textbf{ILFEM}}  &  \makebox[0.12\textwidth][c]{\textbf{CLFEM}}  & \textbf{DSC(\%)}\\
\hline
$\surd$  & & & 79.25\\
$\surd$ & $\surd$ & &79.68\\
$\surd$ &   & $\surd$ & 79.93\\
$\surd$ & $\surd$& $\surd$ & \textbf{80.22} \\ 
\hline
\hline
\end{tabular}
\end{table}

\subsection{Ablation Study}
\vspace{0.1in}
\subsubsection{Model Complexity}
\vspace{0.1in}
   It is well known that heavyweight networks tend to overfit on a small amount of medical image data.  As the results of the experiment shown in Table \ref{tab5}. Compared with TransUNet \cite{chen2021transunet}, SSFormer \cite{wang2022stepwise}, HiFormer \cite{heidari2023hiformer}, SwinUNet \cite{cao2023swin}  which have 105.32M, 66.22M, 29.52M, 27.17M parameters respectively, the DFEN model achieves significant performance improvement  while tending towards lightweight design, with only 24.01M parameters.

\subsubsection{The effectiveness of ILFEM and CLFEM}
\vspace{0.1in}
We conducted ablation experiments on the image-level feature equalization module and the class-level feature equalization module on the BUSI dataset. Table \ref{tab6} showcases the quantitative experimental results, showing that integrating ILFEM into the baseline model leads to a performance improvement of 0.43\%, while integrating CLFEM into the baseline model results in a performance improvement of 0.68\%.  Compared with ILFEM, the reason why CLFEM brings more improvements to the network is that in medical images, the number of pixels in the target area is much smaller than that in the background area. CLFEM can encourage the network to pay more attention to these target areas with fewer pixels, thereby improving more performance. When ILFEM and CLFEM are combined, they collectively enhance the performance of the model by 0.97\%.  These results strongly support the effectiveness of  ILFEM and CLFEM.

\begin{figure*}
      \centering
	\subfloat[Image]{
		\begin{minipage}[t]{0.16\textwidth}
			\centering
			\includegraphics[width=2.75cm,height=2.5cm]{images_benign_8.png}\\
			\vspace{0.03cm}
                         \includegraphics[width=2.75cm,height=2.5cm]{images_benign_162.png}\\
			\vspace{0.03cm}
                \includegraphics[width=2.75cm,height=2.5cm]{images_benign_322.png}\\
                \vspace{0.03cm}
                \includegraphics[width=2.75cm,height=2.5cm]{images_benign_421.png}\\
                \vspace{0.03cm}
		\end{minipage}%
	}%
	\subfloat[Baseline]{
		\begin{minipage}[t]{0.16\textwidth}
			\centering
			\includegraphics[width=2.75cm,height=2.5cm]{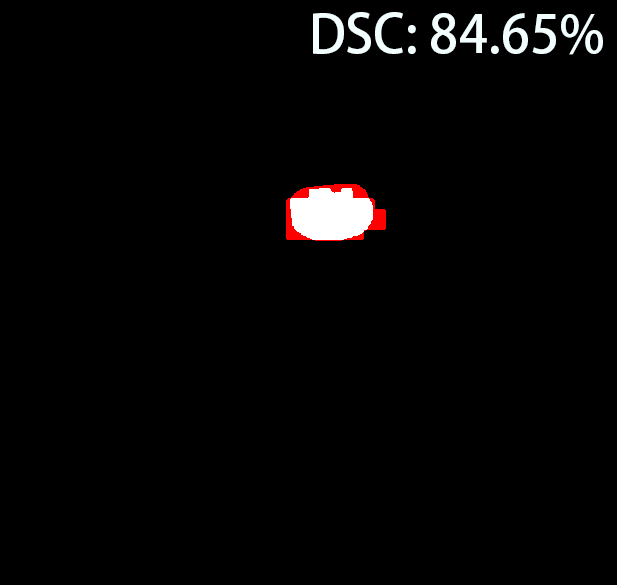}\\
			\vspace{0.03cm}
                   
                        \includegraphics[width=2.75cm,height=2.5cm]{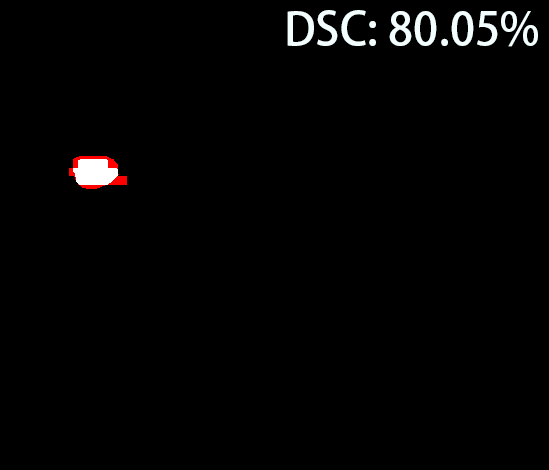}\\
			\vspace{0.03cm}
                        
                \includegraphics[width=2.75cm,height=2.5cm]{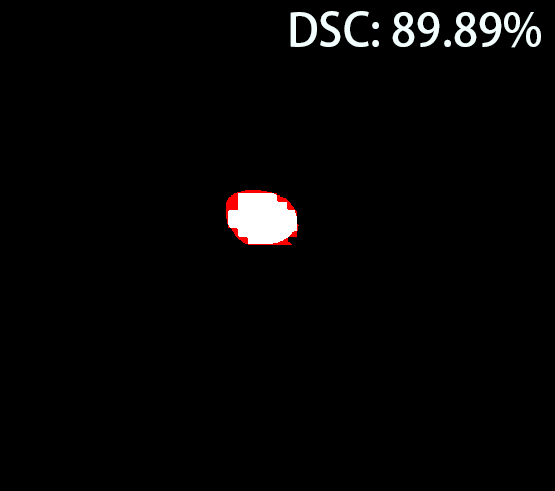}\\
			\vspace{0.03cm}
                 \includegraphics[width=2.75cm,height=2.5cm]{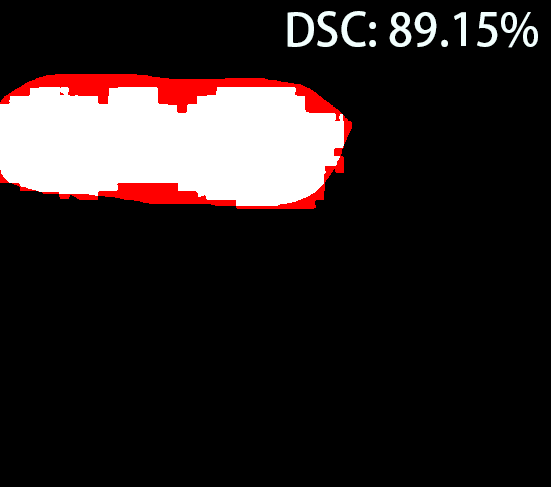}\\
			\vspace{0.03cm}
		\end{minipage}%
	}%
       \subfloat[Baseline + ILFEM]{
		\begin{minipage}[t]{0.16\textwidth}
			\centering
			\includegraphics[width=2.75cm,height=2.5cm]{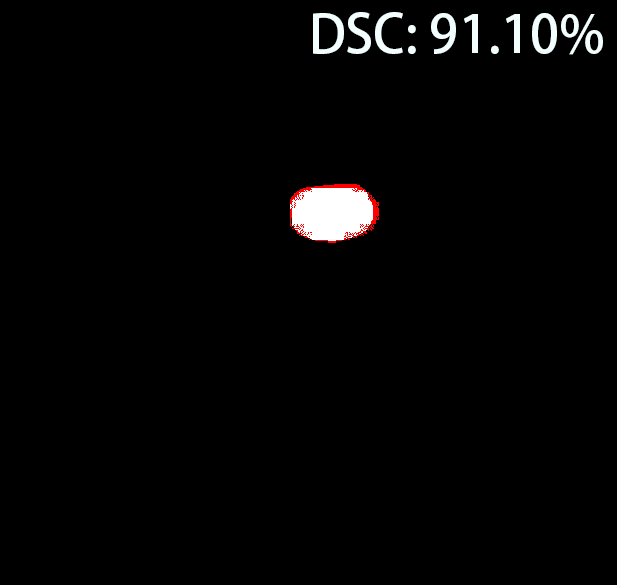}\\
			\vspace{0.03cm}
                    
                        \includegraphics[width=2.75cm,height=2.5cm]{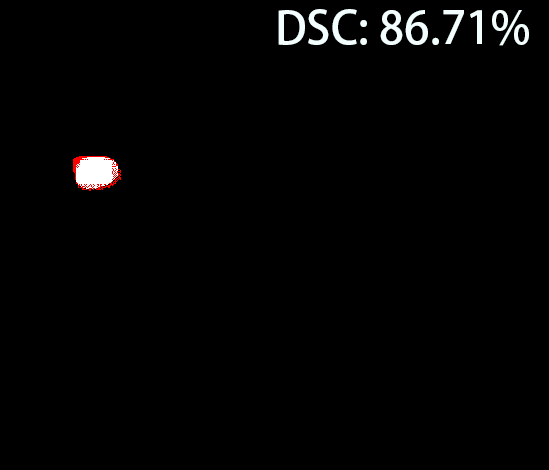}\\
			\vspace{0.03cm}
                        \includegraphics[width=2.75cm,height=2.5cm]{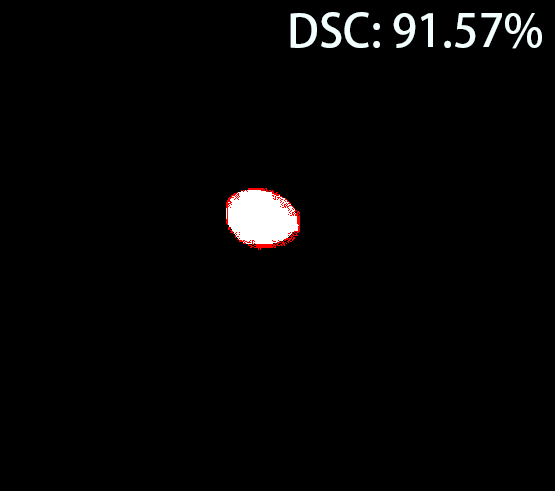}\\
			\vspace{0.03cm}
                \includegraphics[width=2.75cm,height=2.5cm]{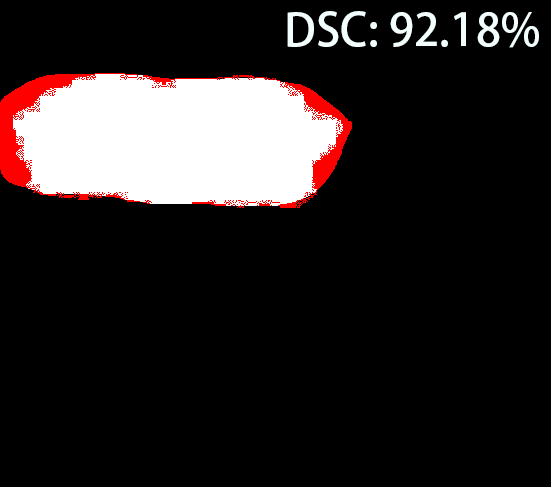}\\
			\vspace{0.03cm}
                    
		\end{minipage}%
	}%
      \subfloat[Baselie+CLFEM]{
		\begin{minipage}[t]{0.16\textwidth}
			\centering
			\includegraphics[width=2.75cm,height=2.5cm]{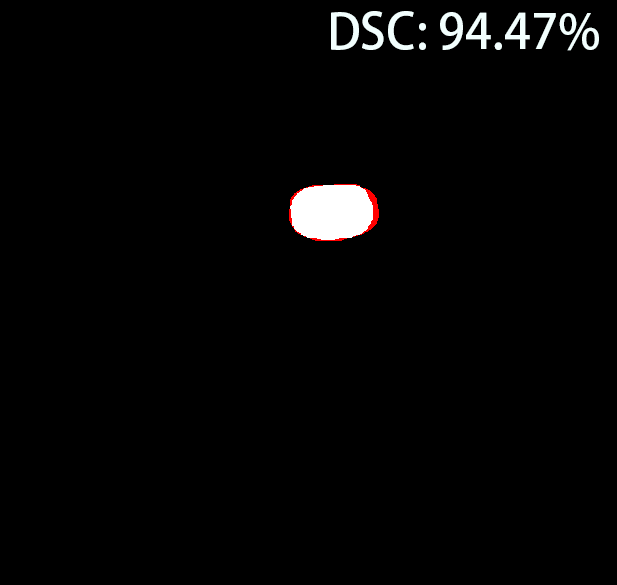}\\
			\vspace{0.03cm}
                   
                        \includegraphics[width=2.75cm,height=2.5cm]{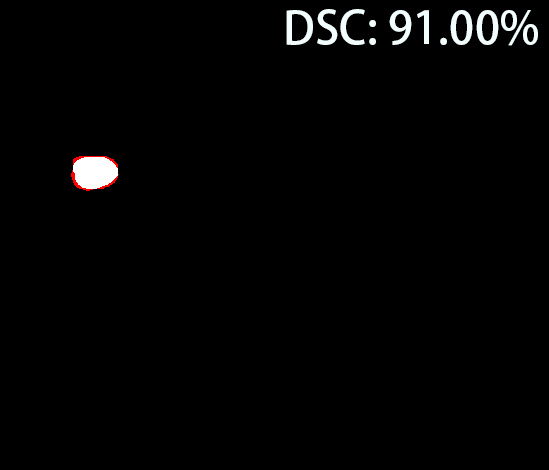}\\
			\vspace{0.03cm}
                        \includegraphics[width=2.75cm,height=2.5cm]{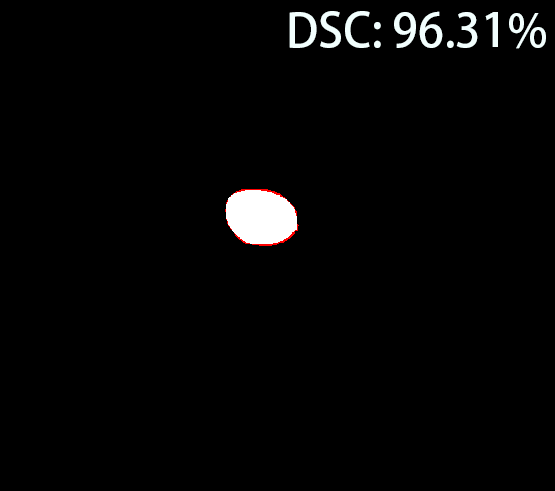}\\
			\vspace{0.03cm}
                \includegraphics[width=2.75cm,height=2.5cm]{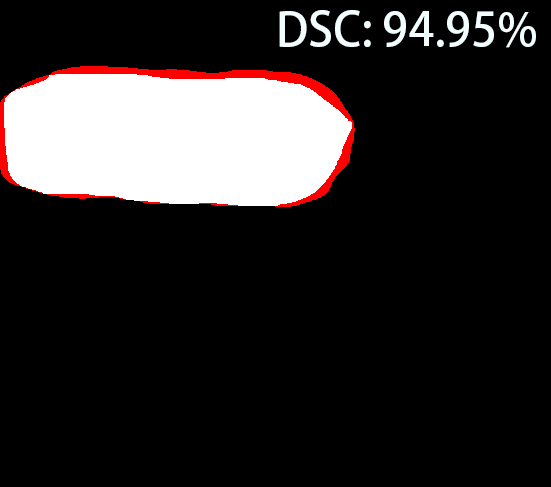}\\
			\vspace{0.03cm}
		\end{minipage}%
	}%
      \subfloat[DFEN]{
		\begin{minipage}[t]{0.16\textwidth}
			\centering
			\includegraphics[width=2.75cm,height=2.5cm]{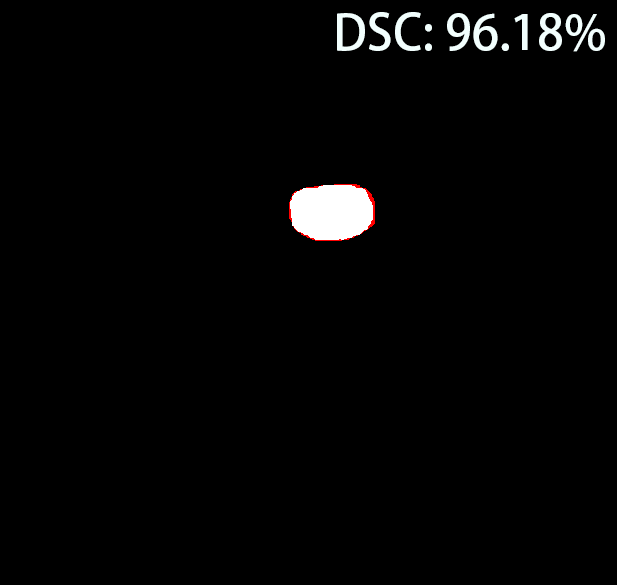}\\
			\vspace{0.03cm}
                     
                        \includegraphics[width=2.75cm,height=2.5cm]{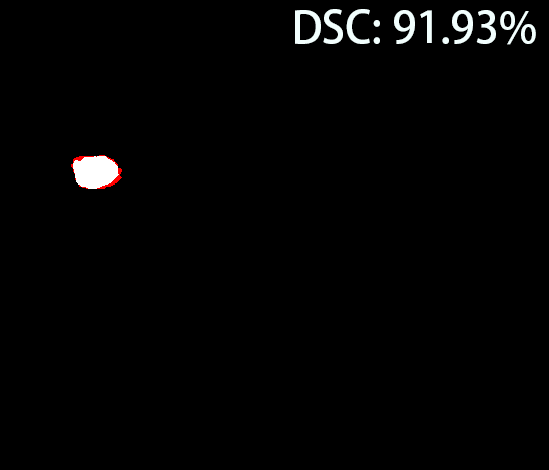}\\
			\vspace{0.03cm}

                        \includegraphics[width=2.75cm,height=2.5cm]{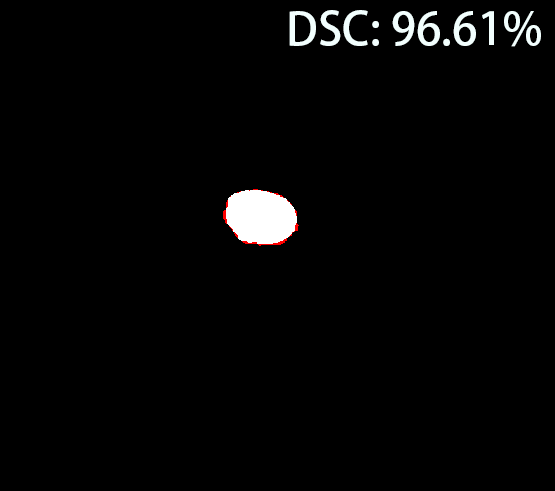}\\
			\vspace{0.03cm}
                \includegraphics[width=2.75cm,height=2.5cm]{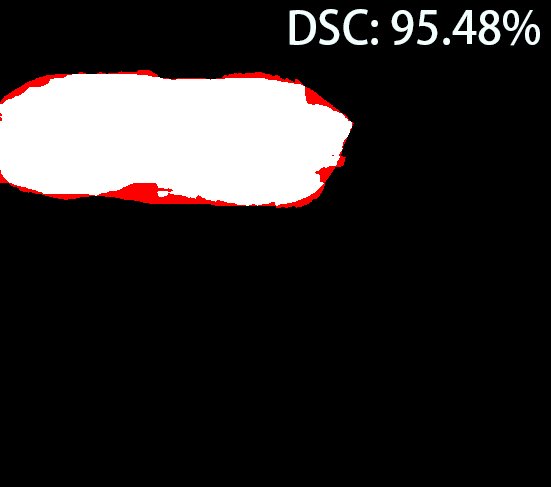}\\
			\vspace{0.03cm}
		\end{minipage}%
	}%
	\subfloat[GT]{
			\begin{minipage}[t]{0.16\textwidth}
				\centering
				\includegraphics[width=2.75cm,height=2.5cm]{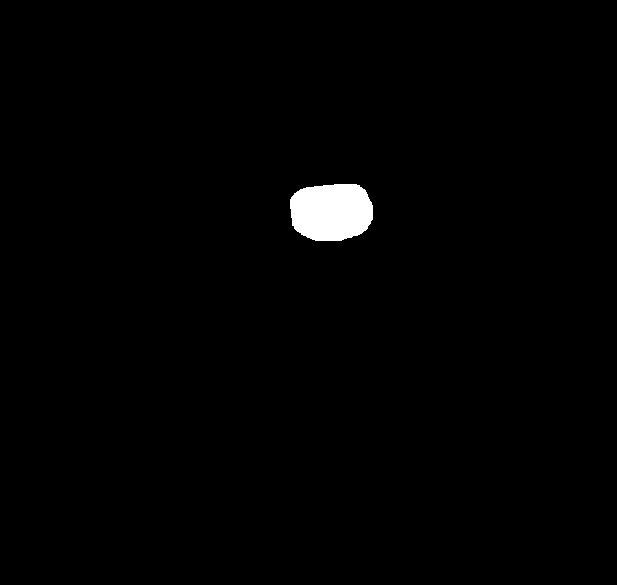}\\
				\vspace{0.03cm}
	                     
	                        \includegraphics[width=2.75cm,height=2.5cm]{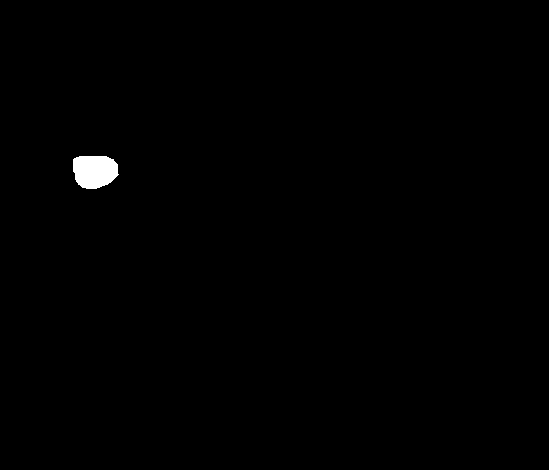}\\
				\vspace{0.03cm}
	
	                        \includegraphics[width=2.75cm,height=2.5cm]{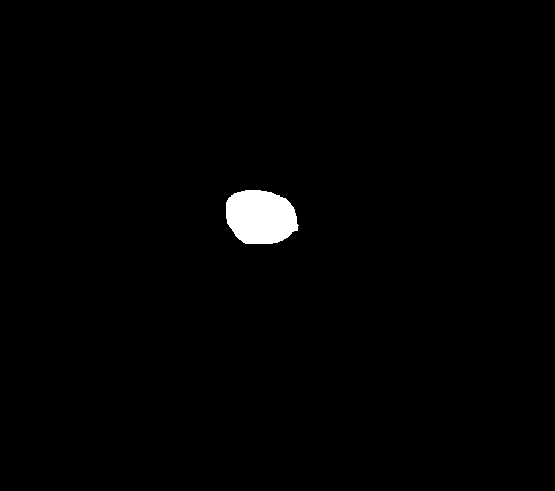}\\
				\vspace{0.03cm}
                     \includegraphics[width=2.75cm,height=2.5cm]{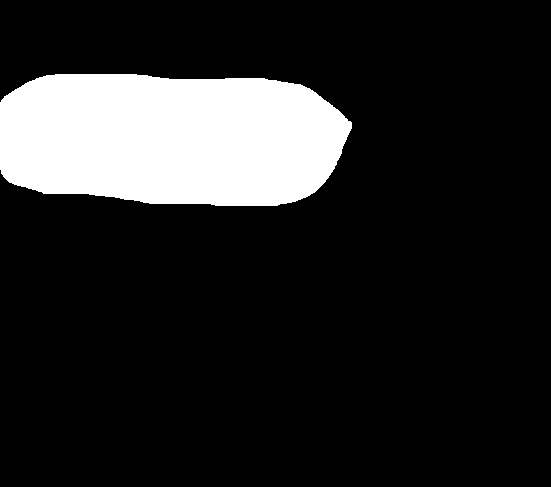}\\
				\vspace{0.03cm}
			\end{minipage}%
		}%
	\caption{Visualization results  to validate the effectiveness of ILFEM and CLFEM. The red region in the figure indicates areas where the model has made classification errors. BaseLine refers to a Transformer model with a pre-trained feature encoder and a  feature decoder.}
	\label{fig7}
\end{figure*}

We conducted visual ablation experiments of ILFEM and CLFEM on the BUSI dataset, with the visualization results depicted in Fig. \ref{fig7}. We employed a Transformer architecture containing only one encoder and one decoder as a baseline. Upon comprehensive examination of both quantitative and visualization experimental results on the BUSI dataset, it can illustrate that our method is effective in alleviating  the  misclassification problem of edge pixels and pixels with a small number of classes due to unequal contextual feature information. Compared with ILFEM, CLFEM exhibits a more pronounced enhancement in model performance. Synergistically incorporating both ILFEM and CLFEM into the baseline  will yield even more substantial performance enhancements.

\begin{table}
\caption{Ablation experiments on $\alpha$ and $\beta$ were conducted on the ISIC2017 testing set. $\alpha$ represents the weight of the cross-entropy loss function, while $\beta$ represents the weight of the Dice loss.}
\label{tab7}
\centering
\resizebox{\linewidth}{!}{
\begin{tabular}{cccccc}
\hline
\hline
\makebox[0.07\textwidth][c]{\textbf{$\alpha$}}  &  \makebox[0.07\textwidth][c]{\textbf{$\beta$}}  & \textbf{DSC(\%)}  & \textbf{SE(\%)} & \textbf{SP(\%)} & \textbf{ACC(\%)}\\
\hline
0.1 & 0.9 & 89.3 & 89.68 & 97.45 & 96.12 \\
0.2 & 0.8 & 90.1 &90.38 & 97.68 & \textbf{96.44} \\
0.3 & 0.7 & \textbf{90.13} & \textbf{91.03} & 97.50 & 96.39 \\
0.4 &0.6 & 90.05 & 90.6 & 97.59 & 96.4 \\
0.5 & 0.5 & 89.7 & 89.96 & 97.60  & 96.29 \\
0.6 & 0.4 & 90.06 & 90.11 & \textbf{97.74} & 96.43 \\
0.7 & 0.3 & 89.95 & 90.48 & 97.57 & 96.36 \\
0.8 & 0.2 & 90.00 & 90.92 & 97.48 & 96.37 \\
0.9 & 0.1 & 89.18 & 89.33 & 97.53 & 91.13\\
\hline
\hline
\end{tabular}
}
\end{table}

\begin{table}
\caption{ Ablation experiments  of  concatenative upsampling strategy on the ACDC testing set. $F_{2,3}$ 
represents the concatenation of additive upsampling features $F_2$ and $F_3$ followed by upsampling.}
\label{tab8}
\centering
\begin{tabular}{cc|ccc}
\hline
\hline
\textbf{Setting}	& \textbf{DSC(\%)} & \textbf{RV} & \textbf{Myo} & \textbf{LV}\\
\hline
$F_3$  & 90.23 & 88.40 & 87.31 & 94.98 \\
$F_{2,3}$ & 90.33 & 88.36 & 87.52 & 95.11 \\
$F_{1,2,3}$ & \textbf{90.46} & \textbf{88.41}  & \textbf{87.81} & \textbf{95.17} \\
\hline
\hline
\end{tabular}
\end{table}

\begin{table}[h]
\setlength{\belowcaptionskip}{0.2cm}
\caption{ Ablation experiments  of  additive upsampling  on the ACDC testing set. Add-up indicates the additive upsampling module.}
\label{tab9}
\centering
\begin{tabular}{ccc|ccc}
\hline
\hline
\textbf{BaseLine} & \textbf{Add-up}	& \textbf{DSC(\%)} & \textbf{RV} & \textbf{Myo} & \textbf{LV}\\
\hline
\usym{1F5F8}  &  & 89.97 & 87.27 & 87.53 & 95.10 \\
\usym{1F5F8} & \usym{1F5F8} & \textbf{90.46} & \textbf{88.41} & \textbf{87.81} & \textbf{95.17} \\
\hline
\hline
\end{tabular}
\end{table}

\subsubsection{The influence of $\alpha$ and $\beta$}
\vspace{0.1in}
We investigated the influence of $\alpha$ and $\beta$ on the ISIC2017 dataset, Table \ref{tab7} displays the specific experimental results.  Cross-entropy loss and Dice loss assist each other in the training of the network according to the experimental results.  The model achieved state-of-the-art result in terms of DSC and SE performance metrics when $\alpha$ and $\beta$ were set to 0.3 and 0.7 respectively. The optimal ACC performance is achieved when $\alpha$ is set to 0.2 and $\beta$ is set to 0.8. The best SP performance is attained when $\alpha$ is set to 0.6 and $\beta$ is set to 0.4. Consequently, we employed the optimal parameter pair of $\alpha$ set to 0.3 and $\beta$ set to 0.7 for all experiments.

\subsubsection{The impact of concatenative upsampling}
\vspace{0.1in}
We performed detailed ablation experiments on concatenative upsampling on the ACDC dataset. Table \ref{tab8} presents the results of the experiments. Based on the experiment results, it is evident that the fused feature obtained by concatenating the three additive upsampling features($F_1$, $F_2$, $F_3$) is more refined, leading to more accurate segmentation results in the RV, Myo, and LV classes. Therefore, We assume that concatenative upsampling was employed in the other experiments conducted in this paper.

\subsubsection{The impact of additive upsampling}
\vspace{0.1in}
Table \ref{tab9} shows the impact of the additive upsampling  on the model performance using the ACDC dataset. BaseLine means that without using skip connection, the enhanced feature representations $R_{aug}$ generated by the dual feature equalization module are upsampled through several patch expanding operations, and then the feature representations generated by each patch expanding are input into concatenative upsampling for feature concatenation. From the experimental results, it can be seen that the additive upsampling  can bring a 0.49\% DSC improvement to the model, thus proving the effectiveness of additive upsampling.

\section{Conclusion}
\vspace{0.1in}
In this paper, we propose a dual feature equalization network for the medical image segmentation  to alleviate the problem of misclassification of pixels caused by border pixels and regions with fewer class pixels capturing more contextual information from other classes.  The image-level feature equalization module is utilized to equalize the contextual information captured by each pixel from other regions, the class-level feature equalization module is employed  to aggregate features from regions of the same class to equalize the feature representations of the corresponding class.  Experimental results on four datasets demonstrate that our method achieves state-of-the-art performance.

 Although our method is effective, there are still some limitations: (1). Our method focuses on scenes with unbalanced number of class pixels, and can achieve state-of-the-art performance in cases of unbalanced class pixels. However, in cases of relatively balanced number of class pixels, our method cannot clearly demonstrate its superiority, and its performance is comparable to the current state-of-the-art methods.
(2). Compared with existing state-of-the-art methods, our model has a low number of parameters and tends to be lightweight, but it still does not enable the model to make real-time predictions that can efficiently assist doctors in diagnosis.
Future work should focus on two aspects. The first is how to improve our method so that the model can further achieve state-of-the-art performance regardless of whether the number of class pixels is unbalanced or balanced.  The second is to minimize the parameters and computation of the model on the basis of improving the performance to achieve real-time medical image segmentation as much as possible.

\section*{Acknowledgements}
This work is supported by the Natural Science Foundation of China (Nos. 62377029, Nos. 92370127 and Nos. 22033002).


\bibliographystyle{elsarticle-num-names}
%\bibliographystyle{cas-model2-names}

\bibliography{reference.bib}

\begin{thebibliography}{63}
\expandafter\ifx\csname natexlab\endcsname\relax\def\natexlab#1{#1}\fi
\providecommand{\url}[1]{\texttt{#1}}
\providecommand{\href}[2]{#2}
\providecommand{\path}[1]{#1}
\providecommand{\DOIprefix}{doi:}
\providecommand{\ArXivprefix}{arXiv:}
\providecommand{\URLprefix}{URL: }
\providecommand{\Pubmedprefix}{pmid:}
\providecommand{\doi}[1]{\href{http://dx.doi.org/#1}{\path{#1}}}
\providecommand{\Pubmed}[1]{\href{pmid:#1}{\path{#1}}}
\providecommand{\bibinfo}[2]{#2}
\ifx\xfnm\relax \def\xfnm[#1]{\unskip,\space#1}\fi
%Type = Inproceedings
\bibitem[{Azad et~al.(2021{\natexlab{a}})Azad, Rouhier, and
  Cohen-Adad}]{azad2021stacked}
\bibinfo{author}{R.~Azad}, \bibinfo{author}{L.~Rouhier},
  \bibinfo{author}{J.~Cohen-Adad},
\newblock \bibinfo{title}{Stacked hourglass network with a multi-level
  attention mechanism: Where to look for intervertebral disc labeling},
\newblock in: \bibinfo{booktitle}{Machine Learning in Medical Imaging: 12th
  International Workshop, MLMI 2021, Held in Conjunction with MICCAI 2021,
  Strasbourg, France, September 27, 2021, Proceedings 12},
  \bibinfo{organization}{Springer}, \bibinfo{year}{2021}{\natexlab{a}}, pp.
  \bibinfo{pages}{406--415}.
%Type = Inproceedings
\bibitem[{Azad et~al.(2021{\natexlab{b}})Azad, Fayjie, Kauffmann, Ben~Ayed,
  Pedersoli, and Dolz}]{azad2021texture}
\bibinfo{author}{R.~Azad}, \bibinfo{author}{A.~R. Fayjie},
  \bibinfo{author}{C.~Kauffmann}, \bibinfo{author}{I.~Ben~Ayed},
  \bibinfo{author}{M.~Pedersoli}, \bibinfo{author}{J.~Dolz},
\newblock \bibinfo{title}{On the texture bias for few-shot cnn segmentation},
\newblock in: \bibinfo{booktitle}{Proceedings of the IEEE/CVF winter conference
  on applications of computer vision}, \bibinfo{year}{2021}{\natexlab{b}}, pp.
  \bibinfo{pages}{2674--2683}.
%Type = Article
\bibitem[{Fu et~al.(2018)Fu, Cheng, Xu, Wong, Liu, and Cao}]{fu2018joint}
\bibinfo{author}{H.~Fu}, \bibinfo{author}{J.~Cheng}, \bibinfo{author}{Y.~Xu},
  \bibinfo{author}{D.~W.~K. Wong}, \bibinfo{author}{J.~Liu},
  \bibinfo{author}{X.~Cao},
\newblock \bibinfo{title}{Joint optic disc and cup segmentation based on
  multi-label deep network and polar transformation},
\newblock \bibinfo{journal}{IEEE transactions on medical imaging}
  \bibinfo{volume}{37} (\bibinfo{year}{2018}) \bibinfo{pages}{1597--1605}.
%Type = Inproceedings
\bibitem[{Kumar et~al.(2018)Kumar, Conjeti, Roy, Wachinger, and
  Navab}]{kumar2018infinet}
\bibinfo{author}{S.~Kumar}, \bibinfo{author}{S.~Conjeti},
  \bibinfo{author}{A.~G. Roy}, \bibinfo{author}{C.~Wachinger},
  \bibinfo{author}{N.~Navab},
\newblock \bibinfo{title}{Infinet: fully convolutional networks for infant
  brain mri segmentation},
\newblock in: \bibinfo{booktitle}{2018 IEEE 15th International Symposium on
  Biomedical Imaging (ISBI 2018)}, \bibinfo{organization}{IEEE},
  \bibinfo{year}{2018}, pp. \bibinfo{pages}{145--148}.
%Type = Inproceedings
\bibitem[{Guo and Yuan(2022)}]{guo2022joint}
\bibinfo{author}{X.~Guo}, \bibinfo{author}{Y.~Yuan},
\newblock \bibinfo{title}{Joint class-affinity loss correction for robust
  medical image segmentation with noisy labels},
\newblock in: \bibinfo{booktitle}{International Conference on Medical Image
  Computing and Computer-Assisted Intervention},
  \bibinfo{organization}{Springer}, \bibinfo{year}{2022}, pp.
  \bibinfo{pages}{588--598}.
%Type = Article
\bibitem[{Ribeiro and Nunes(2023)}]{ribeiro2023left}
\bibinfo{author}{M.~A. Ribeiro}, \bibinfo{author}{F.~L. Nunes},
\newblock \bibinfo{title}{Left ventricle segmentation combining deep learning
  and deformable models with anatomical constraints},
\newblock \bibinfo{journal}{Journal of Biomedical Informatics}
  \bibinfo{volume}{142} (\bibinfo{year}{2023}) \bibinfo{pages}{104366}.
%Type = Article
\bibitem[{Lee et~al.(2001)Lee, Hara, Fujita, Itoh, and
  Ishigaki}]{lee2001automated}
\bibinfo{author}{Y.~Lee}, \bibinfo{author}{T.~Hara},
  \bibinfo{author}{H.~Fujita}, \bibinfo{author}{S.~Itoh},
  \bibinfo{author}{T.~Ishigaki},
\newblock \bibinfo{title}{Automated detection of pulmonary nodules in helical
  ct images based on an improved template-matching technique},
\newblock \bibinfo{journal}{IEEE Transactions on medical imaging}
  \bibinfo{volume}{20} (\bibinfo{year}{2001}) \bibinfo{pages}{595--604}.
%Type = Inproceedings
\bibitem[{Holtzman-Gazit et~al.(2003)Holtzman-Gazit, Goldsher, and
  Kimmel}]{holtzman2003hierarchical}
\bibinfo{author}{M.~Holtzman-Gazit}, \bibinfo{author}{D.~Goldsher},
  \bibinfo{author}{R.~Kimmel},
\newblock \bibinfo{title}{Hierarchical segmentation of thin structures in
  volumetric medical images},
\newblock in: \bibinfo{booktitle}{Medical Image Computing and Computer-Assisted
  Intervention-MICCAI 2003: 6th International Conference, Montr{\'e}al, Canada,
  November 15-18, 2003. Proceedings 6}, \bibinfo{organization}{Springer},
  \bibinfo{year}{2003}, pp. \bibinfo{pages}{562--569}.
%Type = Inproceedings
\bibitem[{Zhan et~al.(2008)Zhan, Zhou, Peng, and Krishnan}]{zhan2008active}
\bibinfo{author}{Y.~Zhan}, \bibinfo{author}{X.~S. Zhou},
  \bibinfo{author}{Z.~Peng}, \bibinfo{author}{A.~Krishnan},
\newblock \bibinfo{title}{Active scheduling of organ detection and segmentation
  in whole-body medical images},
\newblock in: \bibinfo{booktitle}{Medical Image Computing and Computer-Assisted
  Intervention--MICCAI 2008: 11th International Conference, New York, NY, USA,
  September 6-10, 2008, Proceedings, Part I 11},
  \bibinfo{organization}{Springer}, \bibinfo{year}{2008}, pp.
  \bibinfo{pages}{313--321}.
%Type = Article
\bibitem[{Chen et~al.(2022)Chen, Zhou, Zhu, Cao, Gu, and Yu}]{chen2022mtdcnet}
\bibinfo{author}{W.~Chen}, \bibinfo{author}{W.~Zhou}, \bibinfo{author}{L.~Zhu},
  \bibinfo{author}{Y.~Cao}, \bibinfo{author}{H.~Gu}, \bibinfo{author}{B.~Yu},
\newblock \bibinfo{title}{Mtdcnet: A 3d multi-threading dilated convolutional
  network for brain tumor automatic segmentation},
\newblock \bibinfo{journal}{Journal of Biomedical Informatics}
  \bibinfo{volume}{133} (\bibinfo{year}{2022}) \bibinfo{pages}{104173}.
%Type = Article
\bibitem[{Li et~al.(2024)Li, Liu, Shi, Wei, Li, and Xiao}]{li2024ucfiltransnet}
\bibinfo{author}{L.~Li}, \bibinfo{author}{Q.~Liu}, \bibinfo{author}{X.~Shi},
  \bibinfo{author}{Y.~Wei}, \bibinfo{author}{H.~Li}, \bibinfo{author}{H.~Xiao},
\newblock \bibinfo{title}{Ucfiltransnet: Cross-filtering transformer-based
  network for ct image segmentation},
\newblock \bibinfo{journal}{Expert Systems with Applications}
  \bibinfo{volume}{238} (\bibinfo{year}{2024}) \bibinfo{pages}{121717}.
%Type = Article
\bibitem[{Xiao et~al.(2023)Xiao, Li, Liu, Zhu, and
  Zhang}]{xiao2023transformers}
\bibinfo{author}{H.~Xiao}, \bibinfo{author}{L.~Li}, \bibinfo{author}{Q.~Liu},
  \bibinfo{author}{X.~Zhu}, \bibinfo{author}{Q.~Zhang},
\newblock \bibinfo{title}{Transformers in medical image segmentation: A
  review},
\newblock \bibinfo{journal}{Biomedical Signal Processing and Control}
  \bibinfo{volume}{84} (\bibinfo{year}{2023}) \bibinfo{pages}{104791}.
%Type = Inproceedings
\bibitem[{Ronneberger et~al.(2015)Ronneberger, Fischer, and
  Brox}]{ronneberger2015u}
\bibinfo{author}{O.~Ronneberger}, \bibinfo{author}{P.~Fischer},
  \bibinfo{author}{T.~Brox},
\newblock \bibinfo{title}{U-net: Convolutional networks for biomedical image
  segmentation},
\newblock in: \bibinfo{booktitle}{Medical Image Computing and Computer-Assisted
  Intervention--MICCAI 2015: 18th International Conference, Munich, Germany,
  October 5-9, 2015, Proceedings, Part III 18},
  \bibinfo{organization}{Springer}, \bibinfo{year}{2015}, pp.
  \bibinfo{pages}{234--241}.
%Type = Article
\bibitem[{Zhou et~al.(2019)Zhou, Siddiquee, Tajbakhsh, and
  Liang}]{zhou2019unet++}
\bibinfo{author}{Z.~Zhou}, \bibinfo{author}{M.~M.~R. Siddiquee},
  \bibinfo{author}{N.~Tajbakhsh}, \bibinfo{author}{J.~Liang},
\newblock \bibinfo{title}{Unet++: Redesigning skip connections to exploit
  multiscale features in image segmentation},
\newblock \bibinfo{journal}{IEEE transactions on medical imaging}
  \bibinfo{volume}{39} (\bibinfo{year}{2019}) \bibinfo{pages}{1856--1867}.
%Type = Inproceedings
\bibitem[{Gao et~al.(2021)Gao, Zhou, and Metaxas}]{gao2021utnet}
\bibinfo{author}{Y.~Gao}, \bibinfo{author}{M.~Zhou}, \bibinfo{author}{D.~N.
  Metaxas},
\newblock \bibinfo{title}{Utnet: a hybrid transformer architecture for medical
  image segmentation},
\newblock in: \bibinfo{booktitle}{Medical Image Computing and Computer Assisted
  Intervention--MICCAI 2021: 24th International Conference, Strasbourg, France,
  September 27--October 1, 2021, Proceedings, Part III 24},
  \bibinfo{organization}{Springer}, \bibinfo{year}{2021}, pp.
  \bibinfo{pages}{61--71}.
%Type = Inproceedings
\bibitem[{Milletari et~al.(2016)Milletari, Navab, and Ahmadi}]{milletari2016v}
\bibinfo{author}{F.~Milletari}, \bibinfo{author}{N.~Navab},
  \bibinfo{author}{S.-A. Ahmadi},
\newblock \bibinfo{title}{V-net: Fully convolutional neural networks for
  volumetric medical image segmentation},
\newblock in: \bibinfo{booktitle}{2016 fourth international conference on 3D
  vision (3DV)}, \bibinfo{organization}{Ieee}, \bibinfo{year}{2016}, pp.
  \bibinfo{pages}{565--571}.
%Type = Article
\bibitem[{Li et~al.(2018)Li, Chen, Qi, Dou, Fu, and Heng}]{li2018h}
\bibinfo{author}{X.~Li}, \bibinfo{author}{H.~Chen}, \bibinfo{author}{X.~Qi},
  \bibinfo{author}{Q.~Dou}, \bibinfo{author}{C.-W. Fu}, \bibinfo{author}{P.-A.
  Heng},
\newblock \bibinfo{title}{H-denseunet: hybrid densely connected unet for liver
  and tumor segmentation from ct volumes},
\newblock \bibinfo{journal}{IEEE transactions on medical imaging}
  \bibinfo{volume}{37} (\bibinfo{year}{2018}) \bibinfo{pages}{2663--2674}.
%Type = Inproceedings
\bibitem[{Huang et~al.(2020)Huang, Lin, Tong, Hu, Zhang, Iwamoto, Han, Chen,
  and Wu}]{huang2020unet}
\bibinfo{author}{H.~Huang}, \bibinfo{author}{L.~Lin},
  \bibinfo{author}{R.~Tong}, \bibinfo{author}{H.~Hu},
  \bibinfo{author}{Q.~Zhang}, \bibinfo{author}{Y.~Iwamoto},
  \bibinfo{author}{X.~Han}, \bibinfo{author}{Y.-W. Chen},
  \bibinfo{author}{J.~Wu},
\newblock \bibinfo{title}{Unet 3+: A full-scale connected unet for medical
  image segmentation},
\newblock in: \bibinfo{booktitle}{ICASSP 2020-2020 IEEE International
  Conference on Acoustics, Speech and Signal Processing (ICASSP)},
  \bibinfo{organization}{IEEE}, \bibinfo{year}{2020}, pp.
  \bibinfo{pages}{1055--1059}.
%Type = Article
\bibitem[{Han et~al.(2022)Han, Jian, and Wang}]{HAN2022109512}
\bibinfo{author}{Z.~Han}, \bibinfo{author}{M.~Jian}, \bibinfo{author}{G.-G.
  Wang},
\newblock \bibinfo{title}{Convunext: An efficient convolution neural network
  for medical image segmentation},
\newblock \bibinfo{journal}{Knowledge-Based Systems} \bibinfo{volume}{253}
  (\bibinfo{year}{2022}) \bibinfo{pages}{109512}. \URLprefix
  \url{https://www.sciencedirect.com/science/article/pii/S0950705122007572}.
  \DOIprefix\doi{https://doi.org/10.1016/j.knosys.2022.109512}.
%Type = Article
\bibitem[{Chen et~al.(2023)Chen, Li, Zhang, and Dai}]{chen2023rethinking}
\bibinfo{author}{G.~Chen}, \bibinfo{author}{L.~Li}, \bibinfo{author}{J.~Zhang},
  \bibinfo{author}{Y.~Dai},
\newblock \bibinfo{title}{Rethinking the unpretentious u-net for medical
  ultrasound image segmentation},
\newblock \bibinfo{journal}{Pattern Recognition}  (\bibinfo{year}{2023})
  \bibinfo{pages}{109728}.
%Type = Inproceedings
\bibitem[{Devlin et~al.(2019)Devlin, Chang, Lee, and
  Toutanova}]{devlin-etal-2019-bert}
\bibinfo{author}{J.~Devlin}, \bibinfo{author}{M.-W. Chang},
  \bibinfo{author}{K.~Lee}, \bibinfo{author}{K.~Toutanova},
\newblock \bibinfo{title}{{BERT}: Pre-training of deep bidirectional
  transformers for language understanding},
\newblock in: \bibinfo{booktitle}{Proceedings of the 2019 Conference of the
  North {A}merican Chapter of the Association for Computational Linguistics:
  Human Language Technologies, Volume 1 (Long and Short Papers)},
  \bibinfo{publisher}{Association for Computational Linguistics},
  \bibinfo{address}{Minneapolis, Minnesota}, \bibinfo{year}{2019}, pp.
  \bibinfo{pages}{4171--4186}. \URLprefix
  \url{https://aclanthology.org/N19-1423}.
  \DOIprefix\doi{10.18653/v1/N19-1423}.
%Type = Inproceedings
\bibitem[{Dosovitskiy et~al.(2021)Dosovitskiy, Beyer, Kolesnikov, Weissenborn,
  Zhai, Unterthiner, Dehghani, Minderer, Heigold, Gelly, Uszkoreit, and
  Houlsby}]{dosovitskiy2021an}
\bibinfo{author}{A.~Dosovitskiy}, \bibinfo{author}{L.~Beyer},
  \bibinfo{author}{A.~Kolesnikov}, \bibinfo{author}{D.~Weissenborn},
  \bibinfo{author}{X.~Zhai}, \bibinfo{author}{T.~Unterthiner},
  \bibinfo{author}{M.~Dehghani}, \bibinfo{author}{M.~Minderer},
  \bibinfo{author}{G.~Heigold}, \bibinfo{author}{S.~Gelly},
  \bibinfo{author}{J.~Uszkoreit}, \bibinfo{author}{N.~Houlsby},
\newblock \bibinfo{title}{An image is worth 16x16 words: Transformers for image
  recognition at scale},
\newblock in: \bibinfo{booktitle}{International Conference on Learning
  Representations}, \bibinfo{year}{2021}. \URLprefix
  \url{https://openreview.net/forum?id=YicbFdNTTy}.
%Type = Inproceedings
\bibitem[{Liu et~al.(2021)Liu, Lin, Cao, Hu, Wei, Zhang, Lin, and
  Guo}]{liu2021swin}
\bibinfo{author}{Z.~Liu}, \bibinfo{author}{Y.~Lin}, \bibinfo{author}{Y.~Cao},
  \bibinfo{author}{H.~Hu}, \bibinfo{author}{Y.~Wei},
  \bibinfo{author}{Z.~Zhang}, \bibinfo{author}{S.~Lin},
  \bibinfo{author}{B.~Guo},
\newblock \bibinfo{title}{Swin transformer: Hierarchical vision transformer
  using shifted windows},
\newblock in: \bibinfo{booktitle}{Proceedings of the IEEE/CVF international
  conference on computer vision}, \bibinfo{year}{2021}, pp.
  \bibinfo{pages}{10012--10022}.
%Type = Article
\bibitem[{Chen et~al.(2021)Chen, Lu, Yu, Luo, Adeli, Wang, Lu, Yuille, and
  Zhou}]{chen2021transunet}
\bibinfo{author}{J.~Chen}, \bibinfo{author}{Y.~Lu}, \bibinfo{author}{Q.~Yu},
  \bibinfo{author}{X.~Luo}, \bibinfo{author}{E.~Adeli},
  \bibinfo{author}{Y.~Wang}, \bibinfo{author}{L.~Lu}, \bibinfo{author}{A.~L.
  Yuille}, \bibinfo{author}{Y.~Zhou},
\newblock \bibinfo{title}{Transunet: Transformers make strong encoders for
  medical image segmentation},
\newblock \bibinfo{journal}{arXiv preprint arXiv:2102.04306}
  (\bibinfo{year}{2021}).
%Type = Inproceedings
\bibitem[{Cao et~al.(2022)Cao, Wang, Chen, Jiang, Zhang, Tian, and
  Wang}]{cao2023swin}
\bibinfo{author}{H.~Cao}, \bibinfo{author}{Y.~Wang}, \bibinfo{author}{J.~Chen},
  \bibinfo{author}{D.~Jiang}, \bibinfo{author}{X.~Zhang},
  \bibinfo{author}{Q.~Tian}, \bibinfo{author}{M.~Wang},
\newblock \bibinfo{title}{Swin-unet: Unet-like pure transformer for medical
  image segmentation},
\newblock in: \bibinfo{booktitle}{European Conference on Computer Vision},
  \bibinfo{organization}{Springer}, \bibinfo{year}{2022}, pp.
  \bibinfo{pages}{205--218}.
%Type = Inproceedings
\bibitem[{Zhang et~al.(2021)Zhang, Liu, and Hu}]{zhang2021transfuse}
\bibinfo{author}{Y.~Zhang}, \bibinfo{author}{H.~Liu}, \bibinfo{author}{Q.~Hu},
\newblock \bibinfo{title}{Transfuse: Fusing transformers and cnns for medical
  image segmentation},
\newblock in: \bibinfo{booktitle}{Medical Image Computing and Computer Assisted
  Intervention--MICCAI 2021: 24th International Conference, Strasbourg, France,
  September 27--October 1, 2021, Proceedings, Part I 24},
  \bibinfo{organization}{Springer}, \bibinfo{year}{2021}, pp.
  \bibinfo{pages}{14--24}.
%Type = Article
\bibitem[{Lin et~al.(2022)Lin, Chen, Xu, Zhang, Lu, and Zhang}]{9785614}
\bibinfo{author}{A.~Lin}, \bibinfo{author}{B.~Chen}, \bibinfo{author}{J.~Xu},
  \bibinfo{author}{Z.~Zhang}, \bibinfo{author}{G.~Lu},
  \bibinfo{author}{D.~Zhang},
\newblock \bibinfo{title}{Ds-transunet: Dual swin transformer u-net for medical
  image segmentation},
\newblock \bibinfo{journal}{IEEE Transactions on Instrumentation and
  Measurement} \bibinfo{volume}{71} (\bibinfo{year}{2022})
  \bibinfo{pages}{1--15}. \DOIprefix\doi{10.1109/TIM.2022.3178991}.
%Type = Inproceedings
\bibitem[{Heidari et~al.(2023)Heidari, Kazerouni, Soltany, Azad, Aghdam,
  Cohen-Adad, and Merhof}]{heidari2023hiformer}
\bibinfo{author}{M.~Heidari}, \bibinfo{author}{A.~Kazerouni},
  \bibinfo{author}{M.~Soltany}, \bibinfo{author}{R.~Azad},
  \bibinfo{author}{E.~K. Aghdam}, \bibinfo{author}{J.~Cohen-Adad},
  \bibinfo{author}{D.~Merhof},
\newblock \bibinfo{title}{Hiformer: Hierarchical multi-scale representations
  using transformers for medical image segmentation},
\newblock in: \bibinfo{booktitle}{Proceedings of the IEEE/CVF Winter Conference
  on Applications of Computer Vision}, \bibinfo{year}{2023}, pp.
  \bibinfo{pages}{6202--6212}.
%Type = Inproceedings
\bibitem[{Valanarasu and Patel(2022)}]{valanarasu2022unext}
\bibinfo{author}{J.~M.~J. Valanarasu}, \bibinfo{author}{V.~M. Patel},
\newblock \bibinfo{title}{Unext: Mlp-based rapid medical image segmentation
  network},
\newblock in: \bibinfo{booktitle}{Medical Image Computing and Computer Assisted
  Intervention--MICCAI 2022: 25th International Conference, Singapore,
  September 18--22, 2022, Proceedings, Part V},
  \bibinfo{organization}{Springer}, \bibinfo{year}{2022}, pp.
  \bibinfo{pages}{23--33}.
%Type = Inproceedings
\bibitem[{Cai et~al.(2024)Cai, Lai, Wang, Wang, Sun, and Yao}]{cai2024poly}
\bibinfo{author}{X.~Cai}, \bibinfo{author}{Q.~Lai}, \bibinfo{author}{Y.~Wang},
  \bibinfo{author}{W.~Wang}, \bibinfo{author}{Z.~Sun},
  \bibinfo{author}{Y.~Yao},
\newblock \bibinfo{title}{Poly kernel inception network for remote sensing
  detection},
\newblock in: \bibinfo{booktitle}{Proceedings of the IEEE/CVF Conference on
  Computer Vision and Pattern Recognition}, \bibinfo{year}{2024}, pp.
  \bibinfo{pages}{27706--27716}.
%Type = Article
\bibitem[{Yin et~al.(2025)Yin, Chen, Pei, Yao, Nie, and Hua}]{yin2025semi}
\bibinfo{author}{J.~Yin}, \bibinfo{author}{T.~Chen}, \bibinfo{author}{G.~Pei},
  \bibinfo{author}{Y.~Yao}, \bibinfo{author}{L.~Nie}, \bibinfo{author}{X.~Hua},
\newblock \bibinfo{title}{Semi-supervised semantic segmentation with
  multi-constraint consistency learning},
\newblock \bibinfo{journal}{arXiv preprint arXiv:2503.17914}
  (\bibinfo{year}{2025}).
%Type = Article
\bibitem[{Wu et~al.(2025)Wu, Dai, Chen, Huang, Ma, and Xiao}]{wu2025generative}
\bibinfo{author}{W.~Wu}, \bibinfo{author}{T.~Dai}, \bibinfo{author}{Z.~Chen},
  \bibinfo{author}{X.~Huang}, \bibinfo{author}{F.~Ma},
  \bibinfo{author}{J.~Xiao},
\newblock \bibinfo{title}{Generative prompt controlled diffusion for weakly
  supervised semantic segmentation},
\newblock \bibinfo{journal}{Neurocomputing}  (\bibinfo{year}{2025})
  \bibinfo{pages}{130103}.
%Type = Article
\bibitem[{Yin et~al.(2024)Yin, Yan, Chen, Chen, and Yao}]{yin2024class_class}
\bibinfo{author}{J.~Yin}, \bibinfo{author}{S.~Yan}, \bibinfo{author}{T.~Chen},
  \bibinfo{author}{Y.~Chen}, \bibinfo{author}{Y.~Yao},
\newblock \bibinfo{title}{Class probability space regularization for
  semi-supervised semantic segmentation},
\newblock \bibinfo{journal}{Computer Vision and Image Understanding}
  \bibinfo{volume}{249} (\bibinfo{year}{2024}) \bibinfo{pages}{104146}.
%Type = Article
\bibitem[{Wu et~al.(2024)Wu, Qiu, Song, Chen, Huang, Ma, and
  Xiao}]{wu2024image}
\bibinfo{author}{W.~Wu}, \bibinfo{author}{X.~Qiu}, \bibinfo{author}{S.~Song},
  \bibinfo{author}{Z.~Chen}, \bibinfo{author}{X.~Huang},
  \bibinfo{author}{F.~Ma}, \bibinfo{author}{J.~Xiao},
\newblock \bibinfo{title}{Image augmentation agent for weakly supervised
  semantic segmentation},
\newblock \bibinfo{journal}{arXiv preprint arXiv:2412.20439}
  (\bibinfo{year}{2024}).
%Type = Article
\bibitem[{Li et~al.(2025)Li, Liu, Xiao, Zhou, Liu, and Zhang}]{li2025expert}
\bibinfo{author}{L.~Li}, \bibinfo{author}{J.~Liu}, \bibinfo{author}{H.~Xiao},
  \bibinfo{author}{G.~Zhou}, \bibinfo{author}{Q.~Liu},
  \bibinfo{author}{Z.~Zhang},
\newblock \bibinfo{title}{Expert guidance and partially-labeled data
  collaboration for multi-organ segmentation},
\newblock \bibinfo{journal}{Neural Networks} \bibinfo{volume}{187}
  (\bibinfo{year}{2025}) \bibinfo{pages}{107396}.
%Type = Inproceedings
\bibitem[{Chen et~al.(2024)Chen, Jiang, Pei, Sun, Wang, and
  Yao}]{chen2024knowledge}
\bibinfo{author}{T.~Chen}, \bibinfo{author}{X.~Jiang},
  \bibinfo{author}{G.~Pei}, \bibinfo{author}{Z.~Sun},
  \bibinfo{author}{Y.~Wang}, \bibinfo{author}{Y.~Yao},
\newblock \bibinfo{title}{Knowledge transfer with simulated inter-image erasing
  for weakly supervised semantic segmentation},
\newblock in: \bibinfo{booktitle}{European Conference on Computer Vision},
  \bibinfo{organization}{Springer}, \bibinfo{year}{2024}, pp.
  \bibinfo{pages}{441--458}.
%Type = Inproceedings
\bibitem[{Wu et~al.(2025)Wu, Song, Qiu, Huang, Ma, and Xiao}]{wuimgfu}
\bibinfo{author}{W.~Wu}, \bibinfo{author}{S.~Song}, \bibinfo{author}{X.~Qiu},
  \bibinfo{author}{X.~Huang}, \bibinfo{author}{F.~Ma},
  \bibinfo{author}{J.~Xiao},
\newblock \bibinfo{title}{Image fusion for cross-domain sequential
  recommendation},
\newblock in: \bibinfo{booktitle}{Companion Proceedings of the ACM Web
  Conference}, \bibinfo{year}{2025}.
%Type = Inproceedings
\bibitem[{Yin et~al.(2025)Yin, Chen, Zheng, Zhou, and Gu}]{yin2025uncertainty}
\bibinfo{author}{J.~Yin}, \bibinfo{author}{Y.~Chen},
  \bibinfo{author}{Z.~Zheng}, \bibinfo{author}{J.~Zhou},
  \bibinfo{author}{Y.~Gu},
\newblock \bibinfo{title}{Uncertainty-participation context consistency
  learning for semi-supervised semantic segmentation},
\newblock in: \bibinfo{booktitle}{ICASSP 2025-2025 IEEE International
  Conference on Acoustics, Speech and Signal Processing (ICASSP)},
  \bibinfo{organization}{IEEE}, \bibinfo{year}{2025}, pp.
  \bibinfo{pages}{1--5}.
%Type = Inproceedings
\bibitem[{Chen et~al.(2024)Chen, Yao, Huang, Li, Nie, and
  Tang}]{chen2024spatial}
\bibinfo{author}{T.~Chen}, \bibinfo{author}{Y.~Yao},
  \bibinfo{author}{X.~Huang}, \bibinfo{author}{Z.~Li},
  \bibinfo{author}{L.~Nie}, \bibinfo{author}{J.~Tang},
\newblock \bibinfo{title}{Spatial structure constraints for weakly supervised
  semantic segmentation},
\newblock volume~\bibinfo{volume}{33}, \bibinfo{publisher}{IEEE},
  \bibinfo{year}{2024}, pp. \bibinfo{pages}{1136--1148}.
%Type = Inproceedings
\bibitem[{Fan et~al.(2020)Fan, Ji, Zhou, Chen, Fu, Shen, and
  Shao}]{fan2020pranet}
\bibinfo{author}{D.-P. Fan}, \bibinfo{author}{G.-P. Ji},
  \bibinfo{author}{T.~Zhou}, \bibinfo{author}{G.~Chen},
  \bibinfo{author}{H.~Fu}, \bibinfo{author}{J.~Shen},
  \bibinfo{author}{L.~Shao},
\newblock \bibinfo{title}{Pranet: Parallel reverse attention network for polyp
  segmentation},
\newblock in: \bibinfo{booktitle}{International conference on medical image
  computing and computer-assisted intervention},
  \bibinfo{organization}{Springer}, \bibinfo{year}{2020}, pp.
  \bibinfo{pages}{263--273}.
%Type = Inproceedings
\bibitem[{Ruan et~al.(2022)Ruan, Xiang, Xie, Liu, and Fu}]{9995040}
\bibinfo{author}{J.~Ruan}, \bibinfo{author}{S.~Xiang},
  \bibinfo{author}{M.~Xie}, \bibinfo{author}{T.~Liu}, \bibinfo{author}{Y.~Fu},
\newblock \bibinfo{title}{Malunet: A multi-attention and light-weight unet for
  skin lesion segmentation},
\newblock in: \bibinfo{booktitle}{2022 IEEE International Conference on
  Bioinformatics and Biomedicine (BIBM)}, \bibinfo{year}{2022}, pp.
  \bibinfo{pages}{1150--1156}. \DOIprefix\doi{10.1109/BIBM55620.2022.9995040}.
%Type = Article
\bibitem[{Li et~al.(2018)Li, He, Zhou, Yu, Ni, Chen, Wang, and
  Lei}]{li2018dense}
\bibinfo{author}{H.~Li}, \bibinfo{author}{X.~He}, \bibinfo{author}{F.~Zhou},
  \bibinfo{author}{Z.~Yu}, \bibinfo{author}{D.~Ni}, \bibinfo{author}{S.~Chen},
  \bibinfo{author}{T.~Wang}, \bibinfo{author}{B.~Lei},
\newblock \bibinfo{title}{Dense deconvolutional network for skin lesion
  segmentation},
\newblock \bibinfo{journal}{IEEE journal of biomedical and health informatics}
  \bibinfo{volume}{23} (\bibinfo{year}{2018}) \bibinfo{pages}{527--537}.
%Type = Article
\bibitem[{Lei et~al.(2020)Lei, Xia, Jiang, Jiang, Ge, Xu, Qin, Chen, Wang, and
  Wang}]{lei2020skin}
\bibinfo{author}{B.~Lei}, \bibinfo{author}{Z.~Xia}, \bibinfo{author}{F.~Jiang},
  \bibinfo{author}{X.~Jiang}, \bibinfo{author}{Z.~Ge}, \bibinfo{author}{Y.~Xu},
  \bibinfo{author}{J.~Qin}, \bibinfo{author}{S.~Chen},
  \bibinfo{author}{T.~Wang}, \bibinfo{author}{S.~Wang},
\newblock \bibinfo{title}{Skin lesion segmentation via generative adversarial
  networks with dual discriminators},
\newblock \bibinfo{journal}{Medical Image Analysis} \bibinfo{volume}{64}
  (\bibinfo{year}{2020}) \bibinfo{pages}{101716}.
%Type = Inproceedings
\bibitem[{Hatamizadeh et~al.(2022)Hatamizadeh, Tang, Nath, Yang, Myronenko,
  Landman, Roth, and Xu}]{hatamizadeh2022unetr}
\bibinfo{author}{A.~Hatamizadeh}, \bibinfo{author}{Y.~Tang},
  \bibinfo{author}{V.~Nath}, \bibinfo{author}{D.~Yang},
  \bibinfo{author}{A.~Myronenko}, \bibinfo{author}{B.~Landman},
  \bibinfo{author}{H.~R. Roth}, \bibinfo{author}{D.~Xu},
\newblock \bibinfo{title}{Unetr: Transformers for 3d medical image
  segmentation},
\newblock in: \bibinfo{booktitle}{Proceedings of the IEEE/CVF winter conference
  on applications of computer vision}, \bibinfo{year}{2022}, pp.
  \bibinfo{pages}{574--584}.
%Type = Inproceedings
\bibitem[{Wang et~al.(2022)Wang, Huang, Tang, Meng, Su, and
  Song}]{wang2022stepwise}
\bibinfo{author}{J.~Wang}, \bibinfo{author}{Q.~Huang},
  \bibinfo{author}{F.~Tang}, \bibinfo{author}{J.~Meng},
  \bibinfo{author}{J.~Su}, \bibinfo{author}{S.~Song},
\newblock \bibinfo{title}{Stepwise feature fusion: Local guides global},
\newblock in: \bibinfo{booktitle}{International Conference on Medical Image
  Computing and Computer-Assisted Intervention},
  \bibinfo{organization}{Springer}, \bibinfo{year}{2022}, pp.
  \bibinfo{pages}{110--120}.
%Type = Article
\bibitem[{Park and Lee(2022)}]{park2022swine}
\bibinfo{author}{K.-B. Park}, \bibinfo{author}{J.~Y. Lee},
\newblock \bibinfo{title}{Swine-net: Hybrid deep learning approach to novel
  polyp segmentation using convolutional neural network and swin transformer},
\newblock \bibinfo{journal}{Journal of Computational Design and Engineering}
  \bibinfo{volume}{9} (\bibinfo{year}{2022}) \bibinfo{pages}{616--632}.
%Type = Article
\bibitem[{Tang et~al.(2023)Tang, Cheang, Yu, Liang, Feng, and
  Chen}]{TANG2023105131}
\bibinfo{author}{S.~Tang}, \bibinfo{author}{C.~F. Cheang},
  \bibinfo{author}{X.~Yu}, \bibinfo{author}{Y.~Liang},
  \bibinfo{author}{Q.~Feng}, \bibinfo{author}{Z.~Chen},
\newblock \bibinfo{title}{Transcs-net: A hybrid transformer-based
  privacy-protecting network using compressed sensing for medical image
  segmentation},
\newblock \bibinfo{journal}{Biomedical Signal Processing and Control}
  \bibinfo{volume}{86} (\bibinfo{year}{2023}) \bibinfo{pages}{105131}.
  \URLprefix
  \url{https://www.sciencedirect.com/science/article/pii/S1746809423005645}.
  \DOIprefix\doi{https://doi.org/10.1016/j.bspc.2023.105131}.
%Type = Inproceedings
\bibitem[{Tan and Le(2019)}]{tan2019efficientnet}
\bibinfo{author}{M.~Tan}, \bibinfo{author}{Q.~Le},
\newblock \bibinfo{title}{Efficientnet: Rethinking model scaling for
  convolutional neural networks},
\newblock in: \bibinfo{booktitle}{International conference on machine
  learning}, \bibinfo{organization}{PMLR}, \bibinfo{year}{2019}, pp.
  \bibinfo{pages}{6105--6114}.
%Type = Article
\bibitem[{Xu et~al.(2023)Xu, Wang, Wang, and Huang}]{xu2023phcu}
\bibinfo{author}{J.~Xu}, \bibinfo{author}{X.~Wang}, \bibinfo{author}{W.~Wang},
  \bibinfo{author}{W.~Huang},
\newblock \bibinfo{title}{Phcu-net: A parallel hierarchical cascade u-net for
  skin lesion segmentation},
\newblock \bibinfo{journal}{Biomedical Signal Processing and Control}
  \bibinfo{volume}{86} (\bibinfo{year}{2023}) \bibinfo{pages}{105262}.
%Type = Inproceedings
\bibitem[{Hu et~al.(2018)Hu, Gu, Zhang, Dai, and Wei}]{hu2018relation}
\bibinfo{author}{H.~Hu}, \bibinfo{author}{J.~Gu}, \bibinfo{author}{Z.~Zhang},
  \bibinfo{author}{J.~Dai}, \bibinfo{author}{Y.~Wei},
\newblock \bibinfo{title}{Relation networks for object detection},
\newblock in: \bibinfo{booktitle}{Proceedings of the IEEE conference on
  computer vision and pattern recognition}, \bibinfo{year}{2018}, pp.
  \bibinfo{pages}{3588--3597}.
%Type = Inproceedings
\bibitem[{Hu et~al.(2019)Hu, Zhang, Xie, and Lin}]{hu2019local}
\bibinfo{author}{H.~Hu}, \bibinfo{author}{Z.~Zhang}, \bibinfo{author}{Z.~Xie},
  \bibinfo{author}{S.~Lin},
\newblock \bibinfo{title}{Local relation networks for image recognition},
\newblock in: \bibinfo{booktitle}{Proceedings of the IEEE/CVF International
  Conference on Computer Vision}, \bibinfo{year}{2019}, pp.
  \bibinfo{pages}{3464--3473}.
%Type = Article
\bibitem[{Bernard et~al.(2018)Bernard, Lalande, Zotti, Cervenansky, Yang, Heng,
  Cetin, Lekadir, Camara, Ballester et~al.}]{bernard2018deep}
\bibinfo{author}{O.~Bernard}, \bibinfo{author}{A.~Lalande},
  \bibinfo{author}{C.~Zotti}, \bibinfo{author}{F.~Cervenansky},
  \bibinfo{author}{X.~Yang}, \bibinfo{author}{P.-A. Heng},
  \bibinfo{author}{I.~Cetin}, \bibinfo{author}{K.~Lekadir},
  \bibinfo{author}{O.~Camara}, \bibinfo{author}{M.~A.~G. Ballester}, et~al.,
\newblock \bibinfo{title}{Deep learning techniques for automatic mri cardiac
  multi-structures segmentation and diagnosis: is the problem solved?},
\newblock \bibinfo{journal}{IEEE transactions on medical imaging}
  \bibinfo{volume}{37} (\bibinfo{year}{2018}) \bibinfo{pages}{2514--2525}.
%Type = Inproceedings
\bibitem[{Codella et~al.(2018)Codella, Gutman, Celebi, Helba, Marchetti, Dusza,
  Kalloo, Liopyris, Mishra, Kittler et~al.}]{codella2018skin}
\bibinfo{author}{N.~C. Codella}, \bibinfo{author}{D.~Gutman},
  \bibinfo{author}{M.~E. Celebi}, \bibinfo{author}{B.~Helba},
  \bibinfo{author}{M.~A. Marchetti}, \bibinfo{author}{S.~W. Dusza},
  \bibinfo{author}{A.~Kalloo}, \bibinfo{author}{K.~Liopyris},
  \bibinfo{author}{N.~Mishra}, \bibinfo{author}{H.~Kittler}, et~al.,
\newblock \bibinfo{title}{Skin lesion analysis toward melanoma detection: A
  challenge at the 2017 international symposium on biomedical imaging (isbi),
  hosted by the international skin imaging collaboration (isic)},
\newblock in: \bibinfo{booktitle}{2018 IEEE 15th international symposium on
  biomedical imaging (ISBI 2018)}, \bibinfo{organization}{IEEE},
  \bibinfo{year}{2018}, pp. \bibinfo{pages}{168--172}.
%Type = Inproceedings
\bibitem[{Mendon{\c{c}}a et~al.(2013)Mendon{\c{c}}a, Ferreira, Marques, Marcal,
  and Rozeira}]{mendoncca2013ph}
\bibinfo{author}{T.~Mendon{\c{c}}a}, \bibinfo{author}{P.~M. Ferreira},
  \bibinfo{author}{J.~S. Marques}, \bibinfo{author}{A.~R. Marcal},
  \bibinfo{author}{J.~Rozeira},
\newblock \bibinfo{title}{Ph 2-a dermoscopic image database for research and
  benchmarking},
\newblock in: \bibinfo{booktitle}{2013 35th annual international conference of
  the IEEE engineering in medicine and biology society (EMBC)},
  \bibinfo{organization}{IEEE}, \bibinfo{year}{2013}, pp.
  \bibinfo{pages}{5437--5440}.
%Type = Article
\bibitem[{Al-Dhabyani et~al.(2020)Al-Dhabyani, Gomaa, Khaled, and
  Fahmy}]{al2020dataset}
\bibinfo{author}{W.~Al-Dhabyani}, \bibinfo{author}{M.~Gomaa},
  \bibinfo{author}{H.~Khaled}, \bibinfo{author}{A.~Fahmy},
\newblock \bibinfo{title}{Dataset of breast ultrasound images},
\newblock \bibinfo{journal}{Data in brief} \bibinfo{volume}{28}
  (\bibinfo{year}{2020}) \bibinfo{pages}{104863}.
%Type = Article
\bibitem[{Oktay et~al.(2018)Oktay, Schlemper, Folgoc, Lee, Heinrich, Misawa,
  Mori, McDonagh, Hammerla, Kainz et~al.}]{oktay2018attention}
\bibinfo{author}{O.~Oktay}, \bibinfo{author}{J.~Schlemper},
  \bibinfo{author}{L.~L. Folgoc}, \bibinfo{author}{M.~Lee},
  \bibinfo{author}{M.~Heinrich}, \bibinfo{author}{K.~Misawa},
  \bibinfo{author}{K.~Mori}, \bibinfo{author}{S.~McDonagh},
  \bibinfo{author}{N.~Y. Hammerla}, \bibinfo{author}{B.~Kainz}, et~al.,
\newblock \bibinfo{title}{Attention u-net: Learning where to look for the
  pancreas},
\newblock \bibinfo{journal}{arXiv preprint arXiv:1804.03999}
  (\bibinfo{year}{2018}).
%Type = Inproceedings
\bibitem[{Valanarasu et~al.(2021)Valanarasu, Oza, Hacihaliloglu, and
  Patel}]{valanarasu2021medical}
\bibinfo{author}{J.~M.~J. Valanarasu}, \bibinfo{author}{P.~Oza},
  \bibinfo{author}{I.~Hacihaliloglu}, \bibinfo{author}{V.~M. Patel},
\newblock \bibinfo{title}{Medical transformer: Gated axial-attention for
  medical image segmentation},
\newblock in: \bibinfo{booktitle}{Medical Image Computing and Computer Assisted
  Intervention--MICCAI 2021: 24th International Conference, Strasbourg, France,
  September 27--October 1, 2021, Proceedings, Part I 24},
  \bibinfo{organization}{Springer}, \bibinfo{year}{2021}, pp.
  \bibinfo{pages}{36--46}.
%Type = Article
\bibitem[{Al-Masni et~al.(2018)Al-Masni, Al-Antari, Choi, Han, and
  Kim}]{al2018skin}
\bibinfo{author}{M.~A. Al-Masni}, \bibinfo{author}{M.~A. Al-Antari},
  \bibinfo{author}{M.-T. Choi}, \bibinfo{author}{S.-M. Han},
  \bibinfo{author}{T.-S. Kim},
\newblock \bibinfo{title}{Skin lesion segmentation in dermoscopy images via
  deep full resolution convolutional networks},
\newblock \bibinfo{journal}{Computer methods and programs in biomedicine}
  \bibinfo{volume}{162} (\bibinfo{year}{2018}) \bibinfo{pages}{221--231}.
%Type = Article
\bibitem[{Wu et~al.(2022)Wu, Chen, Chen, Wang, Lei, and Wen}]{wu2022fat}
\bibinfo{author}{H.~Wu}, \bibinfo{author}{S.~Chen}, \bibinfo{author}{G.~Chen},
  \bibinfo{author}{W.~Wang}, \bibinfo{author}{B.~Lei},
  \bibinfo{author}{Z.~Wen},
\newblock \bibinfo{title}{Fat-net: Feature adaptive transformers for automated
  skin lesion segmentation},
\newblock \bibinfo{journal}{Medical image analysis} \bibinfo{volume}{76}
  (\bibinfo{year}{2022}) \bibinfo{pages}{102327}.
%Type = Inproceedings
\bibitem[{Rao et~al.(2022)Rao, Zhao, Tang, Zhou, Lim, and
  Lu}]{NEURIPS2022_436d042b}
\bibinfo{author}{Y.~Rao}, \bibinfo{author}{W.~Zhao}, \bibinfo{author}{Y.~Tang},
  \bibinfo{author}{J.~Zhou}, \bibinfo{author}{S.~N. Lim},
  \bibinfo{author}{J.~Lu},
\newblock \bibinfo{title}{Hornet: Efficient high-order spatial interactions
  with recursive gated convolutions},
\newblock in: \bibinfo{editor}{S.~Koyejo}, \bibinfo{editor}{S.~Mohamed},
  \bibinfo{editor}{A.~Agarwal}, \bibinfo{editor}{D.~Belgrave},
  \bibinfo{editor}{K.~Cho}, \bibinfo{editor}{A.~Oh} (Eds.),
  \bibinfo{booktitle}{Advances in Neural Information Processing Systems},
  volume~\bibinfo{volume}{35}, \bibinfo{publisher}{Curran Associates, Inc.},
  \bibinfo{year}{2022}, pp. \bibinfo{pages}{10353--10366}. \URLprefix
  \url{https://proceedings.neurips.cc/paper_files/paper/2022/file/436d042b2dd81214d23ae43eb196b146-Paper-Conference.pdf}.
%Type = Article
\bibitem[{Wu et~al.(2024)Wu, Liang, Huang, Shi, Gu, Zhu, and
  Chang}]{WU2024105517}
\bibinfo{author}{R.~Wu}, \bibinfo{author}{P.~Liang},
  \bibinfo{author}{X.~Huang}, \bibinfo{author}{L.~Shi},
  \bibinfo{author}{Y.~Gu}, \bibinfo{author}{H.~Zhu},
  \bibinfo{author}{Q.~Chang},
\newblock \bibinfo{title}{Mhorunet: High-order spatial interaction unet for
  skin lesion segmentation},
\newblock \bibinfo{journal}{Biomedical Signal Processing and Control}
  \bibinfo{volume}{88} (\bibinfo{year}{2024}) \bibinfo{pages}{105517}.
  \URLprefix
  \url{https://www.sciencedirect.com/science/article/pii/S1746809423009503}.
  \DOIprefix\doi{https://doi.org/10.1016/j.bspc.2023.105517}.
%Type = Article
\bibitem[{Wang et~al.(2023)Wang, Tang, Xiao, Zhou, Fang, and
  Yang}]{wang2023grenet}
\bibinfo{author}{J.~Wang}, \bibinfo{author}{Y.~Tang},
  \bibinfo{author}{Y.~Xiao}, \bibinfo{author}{J.~T. Zhou},
  \bibinfo{author}{Z.~Fang}, \bibinfo{author}{F.~Yang},
\newblock \bibinfo{title}{Grenet: Gradually recurrent network with curriculum
  learning for 2-d medical image segmentation},
\newblock \bibinfo{journal}{IEEE Transactions on Neural Networks and Learning
  Systems}  (\bibinfo{year}{2023}).
%Type = Article
\bibitem[{Zhang et~al.(2018)Zhang, Liu, and Wang}]{zhang2018road}
\bibinfo{author}{Z.~Zhang}, \bibinfo{author}{Q.~Liu},
  \bibinfo{author}{Y.~Wang},
\newblock \bibinfo{title}{Road extraction by deep residual u-net},
\newblock \bibinfo{journal}{IEEE Geoscience and Remote Sensing Letters}
  \bibinfo{volume}{15} (\bibinfo{year}{2018}) \bibinfo{pages}{749--753}.

\end{thebibliography}

% % Biography
% % Here goes the biography details.

% % Here goes the biography details.

\end{document}